\documentclass[10pt,journal,compsoc]{IEEEtran}
\ifCLASSOPTIONcompsoc
\usepackage[nocompress]{cite}
\else
\usepackage{cite}
\fi
\ifCLASSINFOpdf
\else
\fi
	\usepackage{xcolor}
	\usepackage{amsfonts}
	\usepackage{mathrsfs}
	\usepackage{amsfonts}
	\usepackage{amssymb}
	\usepackage{graphicx}
	\usepackage{epsfig}
	\usepackage{epstopdf}
	\usepackage{psfrag}
	\usepackage{mathtools}
	\usepackage{array}
	\usepackage{subfigure}
	\usepackage{float}
	\usepackage{cite,graphicx,amsmath,amssymb,color}
	\usepackage{stmaryrd}
	\usepackage{multirow}
	\usepackage{subfig}
	\usepackage{graphicx,times,amsmath} 
	\usepackage{enumerate}
	\usepackage{makecell}
	
	\usepackage[linesnumbered,ruled,vlined]{algorithm2e}
	\usepackage{lettrine}
	\usepackage{tabu}
	\usepackage{booktabs}
	\usepackage{color}
	\usepackage{bm}
	
\begin{document}
		\bibliographystyle{IEEEtran}
		%
		\title{Timestamp-supervised Wearable-based Activity Segmentation and Recognition with Contrastive Learning and Order-Preserving Optimal Transport}
		%
		%
		%
		%
		
		\author{Songpengcheng Xia,~\IEEEmembership{Graduate Student Member,~IEEE,}
			Lei Chu,~\IEEEmembership{Senior Member,~IEEE,}
			Ling Pei,~\IEEEmembership{Senior Member,~IEEE,}
                Jiarui Yang,
			Wenxian Yu,~\IEEEmembership{Senior Member,~IEEE,}
			Robert C. Qiu,~\IEEEmembership{Fellow,~IEEE,}
			
			\IEEEcompsocitemizethanks{\IEEEcompsocthanksitem The work was supported by the National Natural Science Foundation of China under Grant 62273229.
				\IEEEcompsocthanksitem Songpengcheng Xia, Ling Pei, Jiarui Yang, Wenxian Yu, and Robert C. Qiu are with the Department
				of Electrical and Computer Engineering, Shanghai Jiao Tong University, Shanghai, China. \protect
				E-mail: {songpengchengxia; ling.pei; jr.yang; wxyu; rcqiu}@sjtu.edu.cn. Lei Chu is with the University of Southern California, Los Angeles, CA, USA. 
			\IEEEcompsocthanksitem The corresponding author is Ling Pei.}
			\thanks{Manuscript received XX, 2023; revised XXX, 2023.}
   }
		
		%
		%

	\markboth{Journal of \LaTeX\ Class Files,~Vol.~14, No.~8, August~2015}%
	{Shell \MakeLowercase{\textit{et al.}}: Bare Demo of IEEEtran.cls for Computer Society Journals}
	%



	\IEEEtitleabstractindextext{%
		\begin{abstract}
			Human activity recognition (HAR) with wearables is one of the serviceable technologies in ubiquitous and mobile computing applications. The sliding-window scheme is widely adopted while suffering from the multi-class windows problem. As a result, there is a growing focus on joint segmentation and recognition with deep-learning methods, aiming at simultaneously dealing with HAR and time-series segmentation issues. However, obtaining the full activity annotations of wearable data sequences is resource-intensive or time-consuming, while unsupervised methods yield poor performance. To address these challenges, we propose a novel method for joint activity segmentation and recognition with timestamp supervision, in which only a single annotated sample is needed in each activity segment. However, the limited information of sparse annotations exacerbates the gap between recognition and segmentation tasks, leading to sub-optimal model performance. Therefore, the prototypes are estimated by class-activation maps to form a sample-to-prototype contrast module for well-structured embeddings. Moreover, with the optimal transport theory, our approach generates the sample-level pseudo-labels that take advantage of unlabeled data between timestamp annotations for further performance improvement. Comprehensive experiments on four public HAR datasets demonstrate that our model trained with timestamp supervision is superior to the state-of-the-art weakly-supervised methods and achieves comparable performance to the fully-supervised approaches.
		\end{abstract}
		
		}

	\maketitle

	\IEEEdisplaynontitleabstractindextext

	%
	\IEEEpeerreviewmaketitle

	\section{Introduction}
	
	\IEEEPARstart{T}{emporal} activity recognition with wearables is one of the essential technologies in intelligent sensing and mobile computing, which has been widely applied in many applications, such as smart home and healthcare services \cite{xiao2022self, hao2021invariant, chen2021deep}. The advancement of mobile computing technologies and wearable devices has allowed wearable-based activity recognition to deliver impressive performance. Compared to visual solutions, these wearable-based methodologies ensure user privacy and are unaffected by occlusions \cite{hao2021invariant, yu2021fedhar, lv2018bi}. 
	
	Traditional wearable-based activity recognition methods employ a fixed-size sliding window to segment continuous sensory data into sub-sequences, which are combined with related activity labels to perform downstream tasks in HAR \cite{pei2020mars, yao2017deepsense}. However, these traditional methods suffer from the multi-class windows problem \cite{yao2018efficient, 9772403}, where the performance of the activity recognition model can be significantly affected by the sliding window's length. Since the duration of the activity is uncertain, the activities in the fixed window often contain multiple action categories, introducing errors to the traditional model training. Various studies \cite{ma2019attnsense, yao2017deepsense} employ related ablation experiments to determine the optimal sliding window length for sensory data segmentation. Hence, more recent investigations \cite{9772403, qian2021weakly, yao2018efficient} have pivoted towards joint activity segmentation and recognition method to handle the multi-class windows problem. Our objective aligns with these studies - to jointly segment and recognize activities within sequences by predicting sample-level labels or dense labels.
	
	Despite the impressive performance of these methods in segmentation and recognition, they necessitate sample-level annotations, a labor-intensive and time-consuming process. Thus, our primary concern is to reduce the quantity of training data annotations without compromising the model's performance. To address this challenge, various research has been undertaken in unsupervised, semi-supervised, and weakly supervised settings in both academia and industry.  Notably, unsupervised methods have seen significant advancements, with applications extending to activity segmentation based on wearable sensory data, such as Change Point Detection (CPD) based methods \cite{deryck2021change, tonekaboni2021unsupervised, truong2020selective, xiao2020deepseg}. Despite these unsupervised methods eliminating the need for extensive data labeling, they present challenges due to their subjectivity and environmental dependency, and generally yield poorer segmentation and recognition performance than fully supervised settings.
	
	\begin{figure}[t]
		\centering
		\includegraphics[width=9cm]{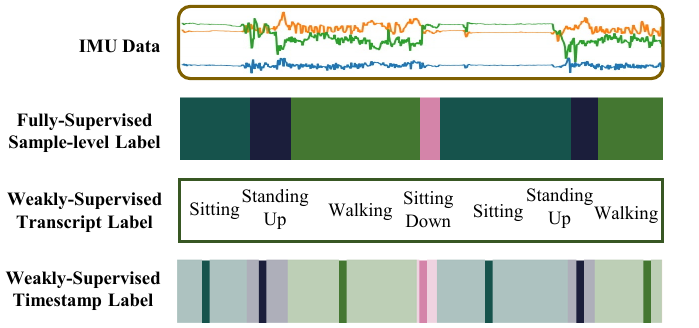}
		\caption{{ Different annotations labels: (a) Fully-supervised setting has the activity label for every sample in the wearable sensory sequence; (b) Transcript-supervised setting contains only the order of activities that occur in the wearable sequence; (c) Timestamp-supervised setting only has the label of a single sample in each activity segment (the wearable sequences usually contain multiple activity segments).}}
		\label{annlevels}
	\end{figure}
	With the recent advancements in semantic segmentation \cite{zhou2022regional, du2022weakly}, the weakly supervised settings have received more attention in recent years due to their potential to balance segmentation performance and annotation effort. Current research on wearable- and video-based activity segmentation includes transcript supervision and timestamp supervision, as shown in Fig.\ref{annlevels}. However, models trained with transcript supervision have shown subpar segmentation and recognition performance, significantly trailing behind fully supervised methods \cite{qian2021weakly, souri2021fast}. Following  \cite{sayed2023new, rahaman2022generalized, khan2022timestamp, li2021temporal}, we adopt a weakly supervised setting with timestamp annotations to train our model, which provides not only the sequence of activities but also partial location information of activities.
	
	For the timestamp supervision, only a single sample from each activity segment is applied to train the segmentation and recognition model \cite{sayed2023new, li2021temporal}. Consequently, the limited information leads to sub-optimal model performance due to the sparsity of the annotations. Therefore, effectively utilizing timestamp labels and numerous unlabeled data becomes critical for training a robust segmentation and recognition model in a weakly supervised setting. To address this challenge, this paper considers it from three aspects:(1) Employing Class Activation Maps (CAMs) allows us to identify sequence locations of existing activity classes that significantly influence activity classification. Leveraging CAMs allows the model to bridge the gap between classification and segmentation, yielding preliminary segmentation results that typically encompass only part of the segmentation area. (2) We aim to develop an effective contrastive learning method that pulls similar samples closer in the feature embedding space while pushing apart the sample features of different activities. Based on the assumption that samples within the same activity category should have similar representations in the feature embedding space, we incorporate a sample-to-prototype contrast module into our model. This approach effectively utilizes the information derived from unlabeled sensory data, fostering more comprehensive and structured feature learning. (3) It is crucial to effectively utilize the information from the activity sequence and timestamp position to generate more plausible pseudo-labels for unlabeled data, which would enhance the training model's accuracy. With the prototype estimated via contrastive learning, our model introduces the optimal transport theory to generate pseudo-labels for further training.
	
	The contributions of our study can be delineated as follows:
	\begin{itemize}
            \item We propose an innovation model for wearable-based joint activity segmentation and recognition with timestamp labels as supervision, which aims to reduce the annotation efforts. This proposed model is trained using only a single annotation in each segment and predicts sample-level activity labels from sensory data.
		
		\item To narrow the gap between the activity recognition and segmentation tasks, our model estimates the prototypes with class-activation maps to form a sample-to-prototype contrast module to obtain a well-structured embedding space.
		
		\item We leverage the optimal transport theory, with an ordering constraint, to compute a probability matrix indicating the association between the samples and the estimated prototypes. With the timestamp locations and labels within the activity segment, our proposed model generates dense pseudo-labels from the optimal transport probability matrix for further improving the segmentation and recognition performance.
		
		\item Comprehensive experiments demonstrate impressive segmentation and recognition performance under both fully supervised and weakly supervised settings, where our model trained with timestamp labels achieves comparable performance to the fully supervised methods. In addition, extensive ablation studies validate the effectiveness of each component within our proposed method.
	\end{itemize}

	\section{Related Work}
	In this section, we review several related works on these four topics: sensor-based activity recognition with sliding window, fully supervised activity segmentation, weakly supervised activity segmentation, and contrastive learning. 
	\subsection{Sensor-based Activity Recognition with Sliding Window}
	Sensor-based HAR has been greatly developed and recognized as a promising technology. Due to the limited information contained in a single sample, the current method segments the sensory data with a fixed-size window and uses the trained neural network to predict the activities, showing impressive recognition performance\cite{pei2020mars, xia2021learning, yao2017deepsense}. Recent developments in deep learning technology have led to the utilization of Deep Neural Networks (DNNs) \cite{radu2018multimodal}, Convolutional Neural Networks (CNNs) \cite{yang2015deep}, Recurrent Neural Networks (RNNs) \cite{guan2017ensembles, chen2019semisupervised}, and hybrid structures \cite{ordonez2016deep} for feature extraction and subsequent activity recognition from wearable data. For example,\cite{yang2015deep} and \cite{ordonez2016deep} employed the fixed sliding window to segment the sensory data and apply the ConvLSTM networks to obtain a promising recognition performance. Ma et al. \cite{ma2019attnsense} designed an activity recognition model based on the attention mechanism and found the optimal window size through experiments. However, due to the uncertainty of the duration of the activity, the multi-class windows problem would bring significant performance degradation to the recognition performance. Therefore, Banos \cite{banos2015multiwindow} and Noor et al. \cite{noor2017adaptive} proposed the adaptive sliding window for these transition activities, where the window size could be adjusted to obtain better recognition performance. These approaches could overcome the multi-class windows problem, while they are most effective on the low-dimensional sensory data. Our method and recent advancing methods choose to directly estimate dense labels for joint activity segmentation and recognition tasks.
	
	\subsection{Fully Supervised Activity Segmentation}
	To handle the multi-class windows problem, an effective solution jointly segments and recognizes activities with the sample-level activity labels\cite{9772403, yao2018efficient}. Inspired by the image semantic segmentation works, \cite{yao2018efficient} and \cite{zhang2019human} respectively applied FCN (Fully Convolutional Networks) and U-net on human activity recognition tasks to predict dense labels, which directly predicts the dense activity labels (sample-level activity labels). Furthermore, the Conditional-UNet with multiple coherent dense labels was proposed by  \cite{zhang2021conditional}, modeling conditional dependency between the multiple labels explicitly. Moreover, our recent work \cite{9772403} advanced the boundary consistency module and the multi-task framework to alleviate the multi-class windows problem and over-segmentation errors, yielding a significant performance improvement. We further  proposed a multi-stage structure with multi-level supervised contrastive loss to learn a well-structured embedding space for enhanced activity segmentation and recognition performance \cite{xia2022multi}. In parallel, action segmentation in video data has garnered considerable interest in both academia and industry \cite{ding2022temporal}. MS-TCN \cite{farha2019ms} is a classic method in video action segmentation tasks, which achieves impressive segmentation results using dilated convolutions with residual connections. On the basis of MS-TCN, \cite{wang2020boundary} and \cite{ishikawa2021alleviating} estimate the action boundary to refine the frame-level label for better segmentation and recognition performance. Moreover, with the development of transformers, Yi et al.\cite{yi2021asformer} proposed ASFormer for temporal action segmentation, which is a transformer-based segmentation model with a multi-stage structure. Predicting sample-level activity labels for joint segmentation and recognition tasks relieves the pain of the multi-class windows issue. However, our goal here is to reduce the workload of data annotation while ensuring segmentation and recognition performance.
	
	\subsection{Weakly Supervised Activity Segmentation}
	Despite the impressive segmentation and recognition results achieved in fully supervised settings, obtaining sample-level annotations is resource-intensive and time-consuming. Therefore, weakly supervised action segmentation has emerged as a viable solution. Qian et al. \cite{qian2021weakly} proposed a unified sensor-based weakly-supervised model for activities segmentation with transcript supervision \cite{bojanowski2014weakly}, which just used the order of activities for model training. Existing weakly supervised temporal action segmentation mainly focuses on image or video data \cite{souri2021fast, rahaman2022generalized, behrmann2022unified}. Based on the Viterbi algorithm and neural network, Richard et al. \cite{richard2018neuralnetwork} proposed the Neural Network Viterbi (NNV), which generates the pseudo ground truth iteratively. Souri et al. \cite{souri2021fast} designed a mutual consistency loss (MuCon) to enforce the frame classification branch and segment generation branch to be consistent. Ridley et al.\cite{ridley2022transformers} proposed a supplemental transcript embedding approach for weakly supervised action segmentation based on transformers. However, using the transcript supervision could reduce the number of annotations but usually causes severe performance degradation\cite{khan2022timestamp}. A novel weakly supervised setting, named timestamp supervision \cite{sayed2023new, behrmann2022unified}, where only one frame is annotated within each activity segment, was first introduced by Li et al. \cite{li2021temporal}. They designed a confidence loss and action detection module to predict frame-level action labels with timestamp annotations. Khan et al. \cite{khan2022timestamp} further enhanced the timestamp supervised activity segmentation method by improving the action boundary detection module and learning a graph convolutional network to generate dense labels from sparse timestamp labels for segmentation models. Behrmann et al.  \cite{behrmann2022unified} incorporated a standard Transformer seq2seq translation model and a novel separate alignment decoder for fully- and timestamp-supervised action segmentation tasks. Rahaman et al. \cite{rahaman2022generalized} proposed an Expectation-Maximization (EM) based generalized and robust framework to accommodate possible annotation errors, e.g., mislabeling. Therefore, weakly supervised settings on various realistic applications can significantly reduce the amount of data labeling. Developing an appropriate weakly supervised model training scheme is critical to balance the annotation effort with segmentation performance. 
	
	\subsection{Contrastive Learning for Time-series Data}
	Contrastive learning \cite{oord2018representation, chen2020simple, zhou2022regional, du2022weakly} has been extensively studied and applied in various research. The main idea of contrastive learning is to learn fair representation in a discriminative manner by using InfoNCE loss or its variants \cite{oord2018representation}. Contrastive predictive coding (CPC) with contrastive loss, proposed by \cite{oord2018representation}, induces the latent space to capture information that is maximally useful to predict future samples. For the computer vision tasks, contrastive learning was usually applied in the self-supervised setting, where multi-view images with augmentation serve as positive samples and negative samples are usually drawn randomly from the datasets \cite{chen2020simple}. Based on the un-/self-supervised learning with contrastive loss, these learned representations generally better generalize to downstream tasks. For time-series data, Deldari et al. \cite{deldari2021time} proposed a change point detection model with self-supervised contrastive predictive coding, which applied contrastive learning on the embeddings of adjacent and separated time-series segments. \cite{eldele2021time} proposed a contextual contrast module with weak and strong augmentations to learn the time-series representation from unlabeled data. Moreover, \cite{khosla2020supervised} extended the self-supervised contrastive method to the fully-supervised setting. Wang et al. \cite{wang2021exploring} applied the supervised contrastive loss to the image semantic segmentation task, which makes the pixel embeddings with the same semantic class more compact and pushes apart the different activity classes. Zhou et al. \cite{zhou2022regional} proposed the regional semantic contrast and aggregation for the dense prediction tasks with weakly supervised annotations. The contrastive learning in weakly supervised settings provides the supervision for unlabeled data, which bridges the gap between classification and segmentation tasks\cite{zhou2022regional, du2022weakly}. Therefore, our proposed method adopts the sample-to-prototype contrast module for further refining the rough activity recognition results (recognition task) in the sequence to the prediction of each sample's activity (segmentation task).
	
	\section{Preliminaries}
	\subsection{Problem Statement}
	We start by defining the $D$-dimensional wearable sensory sequence of length $T$ as $ \bm{X}_{1:T} = [\bm{x}_1,...,\bm{x}_T], \bm{x}_t \in \mathcal{R}^D $, which could be divided in to $N$ segments ($N \ll T$) with the same activity within every segment. Traditional HAR method uses the sliding window to pre-segment the input sequence and predict the activity label of each window, suffering form the multi-class windows problem \cite{9772403,yao2018efficient}. Therefore, the goal of the joint activity recognition and segmentation is to predict the corresponding sample-level activity labels $ \hat{\bm{Y}}_{1:T} = [\hat{\bm{y}}_1,...,\hat{\bm{y}}_T], \hat{\bm{y}}_t \in \mathcal{R}^C $, where $C$ is the number of classes. Most wearable-based activity segmentation is in the fully supervised setting, where the activity labels for every sample $\hat{\bm{y}}_t$ are available in the training phase. However, in our weakly supervised setting with timestamp annotations, only a single sample for each activity segment is annotated. The annotated timestamp labels can be represented by $ \bm{Y}_{TS} = [\bm{y}_{t_1},...,\bm{y}_{t_N}] $ in the training wearable sensory sequence with $T$ samples and $N$ segments, where the sample $t_n$ belongs to the $n$-th activity segment in the training sequence $\bm{X}_{train}$.
	
	In our work, instead of using the fully sample-level activity labels for model training, our goal is to get a robust joint activity segmentation and recognition model with the timestamp labels $\bm{Y}_{TS}$. In the inference stage, our trained model could predict the sample-level labels $\hat{\bm{Y}}$ from the test sequence $\bm{X}_{test}$.

	\subsection{Contrastive Learning with InfoNCE}
	Contrastive learning has been widely investigated in self-supervised learning research, which aims to learn a robust representation by contrasting the anchor ($S$) with positives ($\mathcal{P}_{S}$) and negatives ($\mathcal{N}_{S}$). In a self-supervised manner, the positive sample is usually set as the anchor augmented results, while the negative samples are randomly sampled in the mini-batch. With the theory of mutual information, InfoNCE loss was proposed in Contrastive Predictive Coding (CPC) \cite{oord2018representation}  and widely used in various self-supervised learning tasks, which can be represented as follows:
	\begin{align}
		\label{info}
		\resizebox{.85\linewidth}{!}{$
			L^{NCE} = -\log \frac{\exp{(\boldsymbol{v \cdot v^{+}} / \tau)}}{\exp{(\boldsymbol{v \cdot v^{+}} / \tau)} + \sum\limits_{v^{-} \in \mathcal{N}_S} \exp{(\boldsymbol{v \cdot v^{-}} / \tau)}}
			$},
		\vspace{-0.5cm}
	\end{align}%
	where $\boldsymbol{v}, \boldsymbol{v^+}, \boldsymbol{v^-}$ are the feature embedding of the anchor, the positives and negatives. $\mathcal{N}_S$ is a set of negatives' embedding, and $\tau$ is the temperature hyper-parameter.
	\subsection{Optimal Transport Theory}
	Optimal transport measures the dissimilarity between two probability distributions over a metric space, also called by Wasserstein distance\cite{su2017order}. Conceptually, if we envision each distribution as a unique way of stacking a unit amount of 'dirt' over a space, the Wasserstein distance is the minimum cost required to transform the stack of one distribution into that of the other\cite{su2017order, lim2022order}. Formally, optimal transport theory considers two sets $\bm{A}=(\bm{a}_1,...,\bm{a}_{N_A})$ and $\bm{B}=(\bm{b}_1,...,\bm{b}_{N_B})$ sampled from each distribution, where $ N_A$ and $N_B$ respectively denote the sizes of these sets. To measure the given metric space, the $\bm{\alpha}=(\alpha_1,...,\alpha_{N_A})$ and $\bm{\beta}=(\beta_1,...,\beta_{N_B})$ are the weights on the sampled sets $\bm{A}$ and $\bm{B}$. Without any prior knowledge, these weights can be assumed to be uniformly distributed \cite{su2017order}, leading to $\bm{\alpha}=(\frac{1}{N_A},...,\frac{1}{N_A})$ and $\bm{\beta}=(\frac{1}{N_B},...,\frac{1}{N_B})$. The goal of optimal transport theory is to solve a matrix $\bm{Q} \in \mathcal{R}^{N_A \times N_B}$ of dimension $N_A \times N_B$, which represents the optimal transformation from $\bm{A}$ to $\bm{B}$. Each element $Q_{ij}$ within $\bm{Q}$ could symbolize the quantity of mass transported from $\bm{a}_i$ to $\bm{b}_j$. Following \cite{su2017order}, the set of all the feasible weight matrices is represented by:
	\begin{align}
		\label{TT}
		\mathcal{Q} = \{\bm{Q} \in \mathcal{R}^{N_A \times N_B} | \bm{Q}\bm{1}_{N_B} = \bm{\alpha},\bm{Q}^\mathrm{T}\bm{1}_N = \bm{\beta}\},
	\end{align}%
	where $\bm{1}_{N_A}$ and $\bm{1}_{N_B}$ represent the vectors of ones in dimensions $N_A$ and $N_B$, respectively. With the constraints of Eq.\eqref{TT}, we could solve this optimal transport problem:
	\begin{align}
		\label{maxTR}
		\max_{\bm{Q} \in \mathcal{Q}} Tr(\bm{Q}^\mathrm{T}\bm{A}\bm{B}^\mathrm{T}) + \epsilon H(\bm{Q}),
	\end{align}%
	where $H(\bm{Q}) = -\sum_{i=1}^{N_A} \sum_{j=1}^{N_B} Q_{ij}\log Q_{ij}$ measures the entropy regularization of $\bm{Q}$, $\epsilon$ is the weight of the entropy term. With the iterative Sinkhorn-Knopp algorithm, the optimal transport matrix $\bm{Q}$ could be computed by:
	\begin{align}
		\label{solve_OT}
		\bm{Q}_{OT} = diag(\bm{u})\exp{(\frac{\bm{A}\bm{B}^\mathrm{T}}{\epsilon})}diag(\bm{v}),
	\end{align}%
	where $\bm{u} \in \mathcal{R}^{N_A}$ and $\bm{v} \in \mathcal{R}^{N_B}$ are re-normalization vectors.
		\begin{figure*}[h]
		\centering
		\includegraphics[width=16cm]{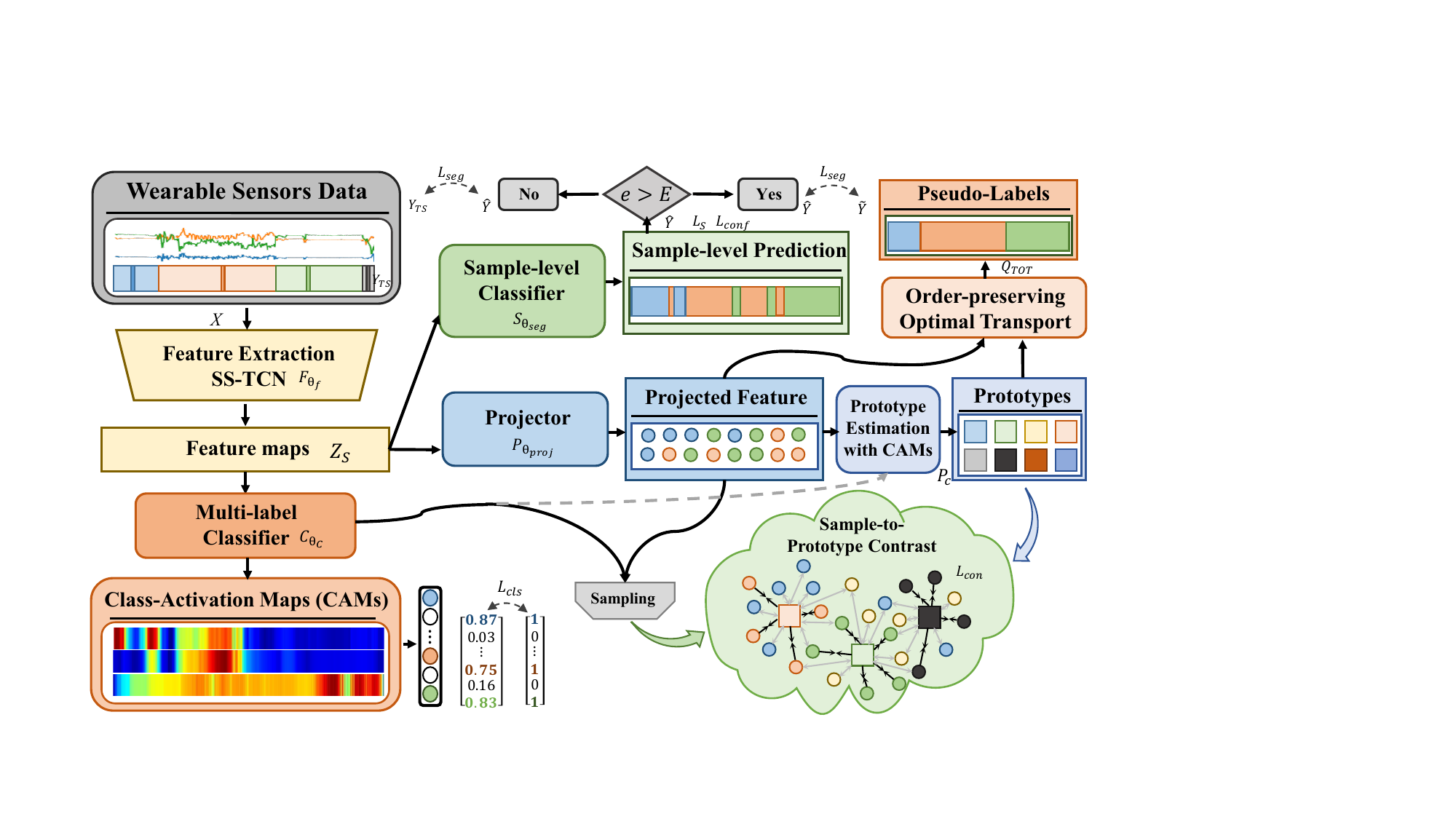}
		\caption{{ Overview of our proposed framework, which consists of four sub-networks. The feature extraction module $F_{\theta_{f}}$ embeds the input wearable data in feature maps $Z_S$. The extracted representations are applied in the multi-label classifier module $C_{\theta_{C}}$, sample-level classifier $F_{\theta_{seg}}$ and projector $F_{\theta_{proj}}$ to get the multi-label classifier's predictions $\hat{Y}_S$, the sample-level activity labels $\hat{Y}$ and the projected feature embeddings $V$. $E$ represents the initial epoch number. When current epoch $e$ is less than $E$, the sample-level classification loss $L_{seg}$ uses the timestamp labels $Y_{TS}$ as the ground truth supervision, and when $e$ is greater than $E$, the generated pseudo-labels $\tilde{Y}$ is used in $L_{seg}$.}}
		\label{overview}
	\end{figure*}
	
	\section{Method}
	\label{method}
	This section introduces the proposed wearable-based weakly activity segmentation and recognition method with sparse timestamp annotations, as shown in Fig. \ref{overview}. Our approach mainly focuses on these challenges: First, we aim to design an effective strategy to address the multi-class windows problem; Second, we strive to utilize the activity class information within each sub-segment efficiently, thereby bridging the gap between the recognition and segmentation tasks; Lastly, we seek to refine the prediction outcomes by integrating the locations of the timestamp annotations within each sub-segment. Therefore, our approach contains three technical modules: 1) a multi-stage temporal convolutional network trained by timestamp annotations for wearable-based activity segmentation and recognition; 2) the sample-to-prototype contrast module with CAMs-constraints prototypes estimation; 3) the sample-level pseudo labels generation module with order-preserving optimal transport theory.

	\subsection{Multi-Stage Temporal Convolutional Network for Timestamp Supervised Activity Segmentation}
	Following the previous work \cite{9772403, yao2018efficient}, to address the multi-class windows problem in the traditional method with fixed-size sliding window, the wearable-based joint activity segmentation and recognition task would be treated as a dense label predicting problem, which uses the network such as FCN to get the sample-level activity predictions. The MS-TCN (Multi-Stage Temporal Convolutional Network) has been proven to be an effective model for extracting contextual information and long-range dependencies within wearable activity sequences \cite{9772403, farha2019ms}. The MS-TCN stacks multiple SS-TCNs, which consist of dilated convolutional layers and residual layers with a dilation factor and $ReLU$ activation. Furthermore, the SS-TCN of the next layer ($l_{th}$) takes the output ($H_{l-1}$) of the previous layer ($l-1_{th}$) as the input to refine the results of the previous layer for better performance, which could be represented by 
	\begin{align}
		\label{tcn}
		\hat{H}_l = ReLU(W_d * H_{l-1} + b_d) \\
		H_l = H_{l-1} + W_r * \hat{H}_l + b_r  ,
	\end{align}%
	where $H_l$ represents the $l_{th}$ layer's output ($H_0$ is the input data), $W_d, W_r$ and $b_d, b_r$ are the weights ans bias vectors of convolutional filters. 
	
	For the timestamp labels $ \bm{Y}_{TS} = [\bm{y}_{t_1},...,\bm{y}_{t_N}] $, one sample is annotated in each activity segment. Our goal is to train a well-performing wearable-based activity segmentation model. Considering the information contained in a single sample ($\bm{x}_t \in \bm{X}$) is limited, our model employs a feature extraction module $F_{\theta_f}(\cdot)$ consisting of the multi-stage TCN network with several dilated residual layers to extract the effective representation ($\bm{Z}_S$) for activity recognition and segmentation, which can be represented by
	\begin{align}
		\label{ZS}
		\bm{Z}_S = F_{\theta_f}(\bm{X}),
	\end{align}
	where the $F_{\theta_f}$ is a shared feature extractor, parameterized by $\theta_f$. Furthermore, a sample-level classifier $S_{\theta_{seg}}$ is employed in our proposed method to get the sample-level activity predictions, while the model is only supervised by sparse timestamp labels. With the extracted representation and the sample-level classifier $S_{\theta_{seg}}$, we can optimize the network's parameters ($\theta_f, \theta_{seg}$) by minimizing
	\begin{align}
		\label{lseg}
		L_{seg} = l(S_{\theta_{seg}}(\bm{Z}_S), \bm{Y}_{TS})=\sum^{N}_{n=1} l(S_{\theta_{seg}}(F_{\theta_f}(\bm{x}_{t_n}), \bm{y}_{t_n}),
	\end{align}
	where $l(\cdot)$ represents the cross-entropy loss, $N$ is the number of segments and valid timestamp labels. It is worth noting that we only use the timestamp label in each activity segment for model training. Furthermore, due to the higher sampling frequency of the IMU and the over-segmentation errors in wearable activity segmentation, a smoothing loss \cite{9772403, farha2019ms, li2021temporal} is employed over the sample-wise log probabilities to smooth the activity predictions:
	\begin{align}
		\label{Lms}
		L_{ms}= - \frac{1}{T_cC} \sum_{t,c} {\widetilde{\Delta^2}_{t,c}}, 
	\end{align}%
	where,
	\begin{equation}
		\begin{aligned}
			{\widetilde{\Delta}_{t,c}} = \left\{\begin{matrix}
                {\Delta_{t,c}}, & {\Delta_{t,c}} \textless \tau\\  
                \tau, & {\Delta_{t,c}} \textgreater \tau
            \end{matrix}\right., \\
			{\Delta_{t,c}} = |log \hat{y}_{t,c} - log \hat{y}_{t-1,c}|,
		\end{aligned}
	\end{equation}
	$C$ represents the activity classes, $T_c$ is the samples' number of $c_{th}$ class, and $ \hat{y}_{t,c} $ represents the probabilities of activity class $c$ at time $t$.
	
	Moreover, the confidence loss proposed by \cite{li2021temporal, khan2022timestamp} is adopted in our proposed method, which encourages our model to predict higher probabilities around the timestamps and forces the predicted probabilities to monotonically decrease as the distance to the timestamps increases. The confidence loss could be represented by 
	\begin{equation}
		\label{Lconf}
		L_{conf}= - \frac{1}{T'} \sum_{\bm{y}_{t_n} \in Y_{TS}} \left({\sum^{t_{n+1}}_{t=t_{n-1}} \delta_{y_{t_n},t}}\right), 
	\end{equation}
	\begin{equation}
		\delta_{y_{t_n},t} = 
		\begin{cases}
			\max(0,\log{\tilde{y}_{t,y_{t_n}}} - \log{\tilde{y}_{t-1,y_{t_n}}})& \text{if $ t \geq t_n $}   \\
			\max(0, \log{\tilde{y}_{t-1,y_{t_n}}} - \log{\tilde{y}_{t,y_{t_n}}})& \text{if $ t < t_n $}
		\end{cases}    
	\end{equation}
	where $\tilde{y}_{t,y_{t_n}}$ is the probability of activity class represented by $y_{t_n}$ at time $t$, and $T' = 2(t_N-t_1)$ is the total number of samples used in this loss. This confidence loss encourages the model to have a high confidence for all samples within an activity segment, and yet it suppresses outlier samples with high confidence that are far from the timestamp annotations \cite{li2021temporal}. It has also been validated in many video action segmentation works \cite{li2021temporal, khan2022timestamp} using timestamp annotations as supervision.
	
	To this end, we obtain the objective for the timestamp supervised setting for wearable-based activity segmentation:
	\begin{align}
		\label{eqxx1}
		\min_{\theta_f,\theta_{seg}} L_{seg} + \lambda_s L_{s} + \lambda_{conf} L_{conf} .
	\end{align}%
	
	The model could be optimized by minimizing Eq.\eqref{eqxx1} with the sparse timestamp labels supervision. However, there are some shortcomings in solving the wearable-based activity segmentation with the weakly supervised setting: 1) simply using the sample-wise cross-entropy loss (with smooth and confidence loss) is not sufficient to bridge the gap between the recognition and segmentation tasks, where the incomplete supervision may lead to poorly structured embeddings \cite{wang2021exploring}; 2) only using the sparse timestamp activity labels for model training ignores the remaining unlabeled samples and the position of the timestamp annotations in the sequence, resulting in inferior recognition and segmentation performance. Therefore, to address these shortcomings, we introduce two core components of our method: 1) we design a sample-to-prototype contrast module with CAMs-constraints prototypes estimation for better sample-level predictions and well-structured embeddings; 2) we propose a sample-level pseudo labels generation module with order-preserving optimal transport theory from sparse timestamp labels for further model training.
	
	\subsection{Sample-to-Prototype Contrast Module with CAMs-constraints Prototypes Estimation}
	The aforementioned objective function (Eq.\eqref{eqxx1}) with sample-level classification loss ($F_{seg}$), smooth loss ($L_S$), confidence loss ($L_{conf}$) and the sparse timestamp labels is hard to meet the need of the segmentation task (or sample-level labels prediction task). Although a sample is labeled in each activity segment, this sample does not necessarily have representative features for this activity class. Inspired by the existing contrastive learning methods \cite{oord2018representation, chen2020simple}, including the fully supervised and unsupervised cases, our proposed method introduces the Class Activation Maps and weakly supervised contrastive loss into the wearable-based activity segmentation and recognition tasks to learn a more structured representation and alleviate the performance degradation caused by incomplete supervision.
	
	\subsubsection{Class Activation Mapping with Multi-label Classifier}
	In the Weakly supervised semantic segmentation task, Class Activation Mapping (CAM) has been proposed to obtain coarse object localization maps from CNN-based image classifiers with global average pooling (GAP), which could overcome the inherent gap between classification and segmentation \cite{zhou2016learning, chen2020weakly}. For a wearable-based task, the feature map can be represented by $\bm{Z}_S \in \mathbb{R}^{D \times T}$, where $D$ is the feature dimension and $T$ is the length of input data. With the GAP operation and the fully connected layer parameterized by $\omega \in \mathbb{R}^{C \times D}$, the CAMs $\bm{m}_c$ for class $c$ can be calculated by:
	\begin{align}
		\label{cams}
		\bm{m}_c = ReLU(\sum^D_{j=1} \omega_{c,j} \bm{z}_{s_{j,:}}),  
	\end{align}%
	where the $\bm{z}_{s_{j,:}} = [z_{s_{j,1}},...,z_{s_{j,T}}]$ represents the feature maps for each dimension. The CAM for a particular class $c$ can indicate the most discriminative segment in the sensory sequence by projecting back the weights of the output layer onto the feature maps\cite{zhou2016learning}. Since there may be more than one activity in the sequence, the classification task is defined as a multi-label classification problem. For the input sequence, we convert the timestamp labels into a sequence-level multi-class classification label $\bm{Y}_S = [y_{s_1},...,y_{s_C}] \in \{0, 1\}^C$ for $C$ classes, where $y_{s_c} = 1$ means the class $c$ exists in the sequence and $y_{s_c} = 0$ otherwise. Based on the extracted feature maps, our method employ the GAP and MLP layers to form the Multi-Label Classifier $C_{\theta_C}(\cdot)$, which could get the classifier's predictions $\hat{\bm{Y}}_S = C_{\theta_C}(\bm{Z}_S)$. With the predictions $\hat{\bm{Y}}_S$ and the sequence-level multi-class ground-truth labels $\bm{Y}_S$, we adopt the multi-label soft margin loss in the training phase:
	\begin{equation}
		\begin{aligned}
			\label{L_cls}
			L_{cls} = l_{cls}(\hat{\bm{Y}}_S, \bm{Y}_S) = -\frac{1}{C-1} \sum^{C-1}_{c=1} [y_{s_c}\log(\frac{1}{1+e^{-\hat{y}_{s_c}}})  \\
			+ (1-y_{s_c})\log(\frac{e^{-\hat{y}_{s_c}}}{1+e^{-\hat{y}_{s_c}}})],
		\end{aligned}
	\end{equation}
	where $l_{cls}(\cdot)$ is the multi-label soft margin loss.
	
	\subsubsection{Prototypes Estimation and Sample-to-Prototype Contrast Module}
	To learn a more structured representation and improve the performance of segmentation, the contrastive loss is additionally applied in our model. Different from our previous work\cite{xia2022multi}, not every sample has a sample-level label and cannot constitute a multi-level contrastive loss (sample-to-segment and segment-to-segment contrast) in weakly supervision. Inspired by \cite{du2022weakly}, we utilize the sample-to-prototype contrast for more structured embedding representations, where the prototypes are estimated by the CAMs and projected feature embeddings. We can use a projection module $P_{\theta_{proj}}$ to get the projected feature embedding $\bm{V} = [\bm{v}_1,...,\bm{v}_T]$, which is represented by:
	\begin{equation}
		\begin{aligned}
			\label{Project}
			\bm{v}_t = P_{\theta_{proj}}(\bm{z}_t)=P_{\theta_{proj}}(F_{\theta_f}(\bm{x}_t)).
		\end{aligned}
	\end{equation}
	
	Since the sparse labeled samples may not be representative of the corresponding category, we use the CAMs to estimate the prototypes $\bm{P}$, where the CAMs represent the confidences that the sample belongs to this category. Therefore, we choose the feature embeddings of the samples with the top $K$ confidences to estimate the prototype $\hat{\bm{P}}_c$ of class $c$:
	\begin{equation}
		\begin{aligned}
			\label{Prototype}
			\hat{\bm{P}}_c = \frac{\sum_{i \in \Omega_c}m_{c,i}\bm{v}_i}{\sum_{i' \in \Omega_c}m_{c,i'}},
		\end{aligned}
	\end{equation}
	where $m_{c,i}$ represents the CAM value of the sample $t$ for class $c$, $i$ belongs to the top $K$ samples' collection, and $\bm{v}_i$ means the projected feature embedding of sample $i$. For storing dataset-level activity information in the whole training phase, we set a global prototype $\bm{P} = \{\bm{P}_1,...,\bm{P}_c,...,\bm{P}_C\}$, which could be updated by the newly estimated prototype $\hat{\bm{P}}$ in each batch. 
	\begin{equation}
		\begin{aligned}
			\label{GP}
			\bm{P}_c = \gamma \hat{\bm{P}}_c + (1-\gamma) \bm{P}_c,
		\end{aligned}
	\end{equation}
	where $\bm{P}_c$ means the global prototype of class $c$, $\hat{\bm{P}}_c$ is the prototype estimated by Eq.\eqref{Prototype} in the current batch, and $\gamma$ is the momentum for prototype update.
	
	With the global prototypes $\bm{P}$, the projected feature embeddings and the pseudo mask ($\bm{y}_p = argmax(\bm{m})$) calculated by CAMs, the sample-to-prototype contrastive loss $L_{con} $ can be represented by:
	\begin{align}
		\label{L_con}
		L_{con} = \frac{1}{\left| \mathcal{I} \right|} \sum\limits_{\bm{v}_t\in \mathcal{I}} log \frac{\exp{(\bm{v}_t \cdot \bm{P}_{y_{p_t}} / \tau)}}{\sum\limits_{\bm{P}_c\in \bm{P}} \exp{(\bm{v}_t \cdot \bm{P}_c / \tau)}},
	\end{align}%
	where $\mathcal{I}$ is the selected samples' feature embeddings set for sample-to-prototype contrastive loss, $y_{p_t} \in [1,2,...,C]$ means the pseudo mask of sample $t$, and the $\bm{P}_{y_{p_t}}$ is the prototype for class $y_{p_t}$ at sample $t$.
	
    Following previous works \cite{kalantidis2020hard, khosla2020supervised, wang2021exploring, du2022weakly}, it is necessary to select the appropriate samples and prototypes for contrastive learning. Therefore, we adopt a novel Timestamp-Constraint Hybrid Hard Example Sampling strategy to choose the prototypes and example samples for sample-to-prototype contrast. For the selection of prototypes in the contrastive loss, our model could get its coarse sample-level labels $y_{p_t}$ from CAMs, which could obtain the positive prototype $\bm{P}_{y_{p_t}}$ and the negative prototypes $\mathcal{P}_N = \mathcal{P}/\bm{P}_{y_{p_t}}$. Following the semi-hard prototype mining approach \cite{wang2021exploring, du2022weakly}, our work hopes to select the 'harder' negative prototypes to participate in sample-to-prototype contrastive learning. The 'harder' negative prototypes are defined as the most similar prototypes in $\mathcal{P}_N$ to the sample's embedding $v_t$, where the 'harder' is measured by whether the dot product of the prototype and sample's embedding is close to 1. Specifically, the negative prototypes first find the top 60\% hardest prototypes in $\mathcal{P}_N$, from which we randomly choose 50\% to form negative sample-prototype pairs in contrastive learning with the selected sample's embedding. For the selection of sensor samples in sample-prototype pairs, we randomly select half of the samples in the wearable sequence and the others are the hard ones, whose dot product result with $\bm{P}_{y_{p_t}}$ is closer to -1 are defined as harder samples. Furthermore, with the timestamp constraint, we can find the samples that are clearly misclassified (e.g. activity predictions between two timestamps do not belong to the category of those two timestamps). Therefore, we additionally form the negative sample pairs between the misclassified samples and the prototype corresponding to the misclassified class, while the prototypes in the positive sample pairs would use a mixture of prototypes corresponding to the classes of the two timestamps. Finally, with sample-to-prototype contrastive loss $L_{con}$ and the multi-label classification loss $L_{cls}$, we attain the objective function as
    \begin{equation}
        \begin{aligned}
		\label{final_sample}
            &\min_{\theta_f,\theta_{seg},\theta_C,\theta_{proj}} L \\
             &= L_{cls} + L_{seg} + \lambda_{con} L_{con} + \lambda_s L_s + \lambda_{conf} L_{conf},
	\end{aligned}
    \end{equation}
	where $\lambda_{con}$, $\lambda_{s}$ and $\lambda_{conf}$ are the hyper-parameters for balancing each loss's contribution.
	
	\subsection{Sample-level Pseudo Label Generation Module with Order-Preserving Optimal Transport Theory}
	Only using the sparse timestamp labels for the activity segmentation model training may lead to inferior performance, which does not take full advantage of a large number of unlabeled samples. Existing action segmentation methods for video sequences\cite{li2021temporal, khan2022timestamp} proposed to generate frame-wise pseudo labels by various strategies, e.g., action change detection\cite{li2021temporal} and graph convolutional network\cite{khan2022timestamp}. However, in wearable-based activity segmentation and recognition tasks, the related pseudo labels generation methods are rarely studied. Considering the balance between training time and model performance, we propose a novel optimal transport based sample-level activity pseudo labels generation method using the timestamp annotations and the prototypes from sample-to-prototype contrast module.
	
	To get the sample-level pseudo labels, our model utilizes the order-preserving optimal transport to compute the optimal transport matrix $\bm{Q}$ with the projected feature embeddings $\bm{V}$ and the prototypes $\bm{P}$, where the element $Q_{ij}$ represents the probability that the projected feature embedding $\bm{v}_i$ map to the prototype $\bm{P}_{c_j}$. It is worth noting that activities usually last for a while, and it is possible to assign adjacent samples to the same prototype as much as possible. Therefore, inspired by \cite{su2017order, kumar2021unsupervised}, based on the prior order-preserving constraints $\bm{T} \in \mathcal{R}^{N \times M}$, we extend Eq.\eqref{solve_OT} with a temporal regularization term, which could preserve the temporal order of the activity:
	\begin{align}
		\label{maxOTT}
		\max_{\bm{Q} \in \mathcal{Q}} Tr(\bm{Q}^TV\bm{P}^T) + \rho KL(\bm{Q}||\bm{T}),
	\end{align}%
	where $KL(\bm{Q}||\bm{T}) = \sum_{i=1}^N \sum_{j=1}^M Q_{ij} \log\frac{Q_{ij}}{T_{ij}}$ is the Kullback-Leibler (KL) divergence between $\bm{Q}$ and $\bm{T}$, $\rho$ is the weight for the KL term, and $\bm{T}$ is modeled by a two-dimensional distribution \cite{su2017order, kumar2021unsupervised}, whose marginal distribution along any line perpendicular to the diagonal is a Gaussian distribution centered at the intersection on the diagonal:
	\begin{align}
		\label{maxOTT}
		T_{ij} = \frac{1}{\sigma \sqrt{2\pi}} \exp(-\frac{d^2_{ij}}{2\sigma^2}), d_{ij} = \frac{|i/N - j/M|}{\sqrt{1/N^2 + 1/M^2}},
	\end{align}%
	where $d_{ij}$ is the distance from the position $(i,j)$ to the diagonal line. To solve the optimal transport problem with order constraints, we obtain the optimal transport matrix $\bm{Q}_{TOT}$ represented by:
	\begin{align}
		\label{solve_TOT}
		\bm{Q}_{TOT} = diag(\bm{u})\exp{(\frac{V\bm{P}^T + \rho\log \bm{T}}{\rho})}diag(\bm{v}).
	\end{align}%
	
	The optimal transport matrix $\bm{Q}_{TOT}$ represents the probability from $\bm{v}_{i}$ mapping to the $\bm{P}_{j}$, where $j \in [1,2,...,C]$. 
	For our timestamp supervised setting, the category of every sample within an activity segment between two timestamps $t_n$ and $t_{n+1}$ can only fall into two possibilities: either it aligns with the previous label $\bm{y}_{t_n}$ or it conforms to the category $\bm{y}_{t_{n+1}}$ of the subsequent timestamp. Consequently, leveraging the estimated optimal transport matrix $\bm{Q}_{TOT}$, our module extracts the probability that each sample belongs to the corresponding prototype ($\bm{P}_{y_{t_n}}$ and $\bm{P}_{y_{t_{n+1}}}$) of the two categories present within this activity segment. A simple approach is to compare these two probabilities that each sample belongs to two prototypes, and assign the activity class with higher probability as a pseudo-label to the unlabeled sample. Despite the order constraints in computing $\bm{Q}_{TOT}$, this simple method of pseudo-label generation from $\bm{Q}_{TOT}$ encounters two key challenges: (1) the pseudo-labels frequently oscillate between two activity categories, contradicting the premise that there should only be one activity change between two timestamps; (2) there tends to be ambiguity and confusion with activities, especially at the boundaries of activity transitions.
	\begin{figure}[t]
		\centering
		\includegraphics[width=9cm]{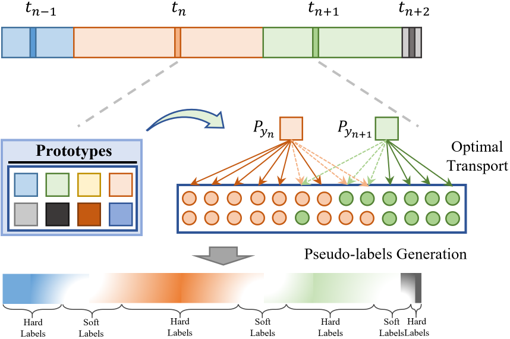}
		\caption{{Sample-level Pseudo-Labels Generation with timestamp annotations and order-preserving optimal transport theory}}
		\label{psegen}
	\end{figure}
	To handle these challenges, we propose a hybrid pseudo-label generation strategy of hard and soft labels. In each activity segment between $t_n$ and $t_{n+1}$, the number of the two activity classes is counted according to the estimation of $\bm{Q}_{TOT}$, which could be represented by $N_{y_{t_n}}$ and $N_{y_{t_{n+1}}}$. Therefore, the final pseudo-labels $\tilde{\bm{Y}}=\{\tilde{\bm{y}}_1,...,\tilde{\bm{y}}_T\}, \tilde{\bm{y}}_t \in \mathcal{R}^C$ can be calculated by:
	\begin{equation}
		\tilde{\bm{y}}_t = 
            \left\{\begin{matrix}
                \bm{y}_{t_n}, & t < (t_n + \varepsilon N_{y_{t_n}})\\  
                \bm{y}_{t_{n+1}}, & t > (t_{n+1} - \varepsilon N_{y_{t_{n+1}}})\\
                \hat{Q}_{i=t,j=\{\bm{P}_{y_{t_n}},\bm{P}_{y_{t_{n+1}}}\}}, & others
            \end{matrix}\right.   
		\label{Y--}
	\end{equation}
	where $\varepsilon$ is the hard label scale parameter, $\hat{Q}_{i=t,j=\{P_{y_{t_n}},P_{y_{t_{n+1}}}\}}$ is the normalized optimal transport matrix making
        \begin{equation}
		\hat{Q}_{i=t,j=P_{y_{t_n}}}+\hat{Q}_{i=t,j=P_{y_{t_{n+1}}}}=1.
		\label{Y--}
	\end{equation}
 
    Based on the segmentation predictions $\hat{\bm{Y}}$ and the pseudo-labels $\tilde{\bm{Y}}$, the sample-level classification loss for all samples could be represented by:
	\begin{align}
		\label{lsega}
		L_{segall} = l_s(F_{\theta_{seg}}(\bm{Z}_S), \tilde{\bm{Y}}) = l_s(\hat{\bm{Y}}, \tilde{\bm{Y}}),
	\end{align}
	\begin{align}
		\label{ls}
		l_s(\hat{\bm{Y}},  \tilde{\bm{Y}}) = -\frac{1}{T} \sum^T_{t=1}\sum^C_{c=1} \tilde{y}_{tc}\log(\hat{y}_{tc}),
	\end{align}
	where $l_s(\cdot)$ is the soft cross-entropy, $T$ is the length of input sequence, and $C$ represents the class number of dataset.
	
	With the above technical analysis, we obtain the final loss function $L$ for weakly supervised activity and segmentation task:
	\begin{equation}
            \begin{aligned}
		\label{final_loss}
		&\min_{\theta_f,\theta_{seg},\theta_C,\theta_{proj}} L \\
         &= L_{cls} + L_{segall} + \lambda_{con} L_{con} + \lambda_s L_s + \lambda_{conf} L_{conf},
	       \end{aligned}
        \end{equation}
	where $\lambda_{con}$, $\lambda_{s}$ and $\lambda_{conf}$ are the hyper-parameters for balancing each loss's contribution. We show the flow of our proposed method in Algorithm \ref{alg1}.
	
	\begin{algorithm}[!t]
		\DontPrintSemicolon
		\KwIn{Training dataset: $\{\bm{X}_{train}, \bm{Y}_{TS_{train}}\}$; Evaluation dataset: $\{\bm{X}_{evl}, \bm{Y}_{evl}\}$}
		\KwOut{Sample-level activity predictions: $\hat{\bm{Y}}$}
		Initialize the parameters $ \theta_{f}, \theta_{seg}, \theta_{C}, \theta_{proj}$; \\
		$e \gets 0, lr \gets 0.001, factor = 0.99 $\;
		\While{$e \leq E^{max}_{epoch}$ or {Convergent}}{
			Get a mini-batch data in training-set;\\  
			\If{$e \leq E^{init}_{epoch}$}{
				Calculate $ L_{cls}, L_{seg}, L_{con}, L_s$ and $L_{conf} $ with Eq.\eqref{L_cls}, \eqref{lseg}, \eqref{L_con}, \eqref{ls} and \eqref{Lconf};\\
				$e \gets e + 1$\;
				Update $ \theta_{f}, \theta_{seg}, \theta_{C}, \theta_{proj} $ based on Eq.\eqref{final_sample}; \\
				\If{mod(e, 100) = 0}{$lr = lr*factor$}
			}
			\If{$e > E^{init}_{epoch}$}{
				Generate the pseudo-labels $\tilde{\bm{Y}}$ from the optimal transport matrix $\bm{Q}_{TOT}$ with Eq.\eqref{solve_TOT}, \eqref{Y--};  \\
				Calculate $ L_{cls}, L_{seg}, L_{con}, L_s$ and $L_{conf} $ with Eq.\eqref{L_cls}, \eqref{lseg}, \eqref{L_con}, \eqref{ls} and \eqref{Lconf};\\
				$e \gets e + 1$\;
				Update $ \theta_{f}, \theta_{seg}, \theta_{C}, \theta_{proj} $ based on Eq.\eqref{final_loss}; \\
				\If{mod(e, 100) = 0}{$lr = lr*factor$}
			}
			
			Calculate $\hat{\bm{Y}}_{evl}$ with Eq.\eqref{lseg} and check convergence;
		}
		
		\Return{Optimized $\theta_{f}, \theta_{seg}, \theta_{C}, \theta_{proj}$}
		
		Predict $\hat{\bm{Y}}$ from test sequence $\bm{X}_{test}$ based on the $S_{\theta_{seg}}(F_{\theta_f}(\bm{X}_{test})$; 
		
		\caption{{\sc The Proposed Algorithm}}
		\label{alg1}
	\end{algorithm}
	
	\section{Experiment}
	In this section, we first introduce the datasets, recognition and segmentation metrics and the competing algorithms used in our experiment. Then, we verify the effectiveness of each module of the model we proposed through various quantitative and qualitative experiments, and compare the state-of-the-art algorithm with significant advantages in recognition and segmentation performance. 
	
	For the experimental setup, we implemented our model on the Pytorch platform with an RTX 3090 GPU. And the Adam optimizer with a learning rate of 0.001 is used in our model's training phase. Following our previous works \cite{9772403, xia2022multi}, we use the MS-TCN \cite{farha2019ms} with ten layers as the backbone. 
	
	\subsection{Evaluation Protocols and Datasets}
	In our experiment, we quantitatively evaluate the model from two aspects of segmentation and recognition. We employ the accuracy ($Acc$), class-average F-score ($F_m$) as the recognition metrics and the Jaccard Index ($JI$) \cite{gammulle2021tmmf}, Intersection over Union ($IoU$) \cite{souri2021fast}, and Overfill/Underfill ($O/U$) \cite{abedin2021attend} as the segmentation metrics.
	
	\subsubsection{Recognition Metrics}
	$\textbf{Accuracy}$: $Acc$ is the overall accuracy for all classes calculated as 
	\begin{equation}
		\begin{split}
			Accuracy = \frac{\sum_{c=1}^{C}TP_{c}+\sum_{c=1}^{C}TN_{c}}{\splitfrac{\sum_{c=1}^{C}TP_{c}+\sum_{c=1}^{C}TN_{c}+}{\sum_{c=1}^{C}FP_{c}+\sum_{c=1}^{C}FN_{c}}},
		\end{split}
	\end{equation}
	where $ C $ denotes the class number, and $TP_{c}$, $FP_{c}$, $TN_{c}$, $FN_{c}$ are the true positives, false positives, true negatives and false negatives of the class $c$, respectively.
	
	\textbf{ Class-average F-score }:$F_m$ reflects the performance of every activity class regardless of its prevalence:
	
	\begin{equation}
		F_m = \frac{2}{C} \sum_{c=1}^{C} \frac{prec_c \cdot recall_c}{prec_c + recall_c},
	\end{equation}
	where $C$ denotes the number of activity classes, $prec_c = TP_c / (TP_c + FP_c)$ and $recall_c = TP_c / (TP_c + FN_c)$ respectively represent the precision and recall terms. 
	
	\subsubsection{Segmentation Metrics}
	
	\textbf{Intersection over Union}: $IoU$ is a popular evaluation metric used in tasks such as image segmentation, which is a measure of the overlap between the predicted and ground truth segments \cite{souri2021fast}. The $IoU$ is computed by the average intersection of a ground truth segment $S$ with the predicted segment $S'$ divided by the union of them, which could be represented by: 
	\begin{equation}
		IoU = \frac{S \cap S' }{S \cup S'},
	\end{equation}
	
	\textbf{Jaccard Index}: $JI$ measures overlap between the ground truth and predicted segments, which is sensitive to over-segmentation errors. Following \cite{gammulle2021tmmf}, we use the $G_{c}$ and $J_{c}$ to represent the ground-truth and predictions of the $c^{th}$ class to get the class-average Jaccard Index:
	\begin{equation}
		JI = \frac{1}{C} \sum_{c=1}^{C}\frac{G_c \cap J_c }{G_c \cup J_c},
	\end{equation}
	where $C$ is the number of classes in this dataset.
	
	\textbf{Overfill and Underfill}: $O/U$ is a commonly used measure in time series segmentation, which indicates the errors when the start and end time of the model's activity predictions are earlier or later than the ones of the ground-truth \cite{ward2011performance}. The $O/U$ could be represented by
        \begin{equation}
		O/U = O + U = O^{\alpha} + O^{\omega} + U^{\alpha} + U^{\omega},
	\end{equation}
        where $O^{\alpha}$ and $O^{\omega}$ mean the $FP$ (False Positives) that occurs at the start or end of a partially matched return, $U^{\alpha}$ and $U^{\omega}$ are the $FN$ (False Negatives) that occurs at the start or end of a detected event.
 
	\subsubsection{Datasets}
	
	\begin{table*}[]
		\centering
		\caption{\sc The Ablation Experiment for the Impact of Loss Formulation on Four Public Datasets}
		\begin{tabular}{c|c|cc|cc|cc|cc}
			\toprule
			\multicolumn{1}{c|}{\multirow{2}{*}{No.}} &\multicolumn{1}{c|}{\multirow{2}{*}{Loss Formulation}} & \multicolumn{2}{c|}{\bfseries{Hospital}} & \multicolumn{2}{c|}{\bfseries{Opportunity}} & \multicolumn{2}{c|}{\bfseries{PAMAP2}}  & \multicolumn{2}{c}{\bfseries{Skoda}}\\ \cmidrule{3-4}  \cmidrule{5-6}  \cmidrule{7-8} \cmidrule{9-10} 
			& & $F_m$  & $Acc$   & $F_m$  & $Acc$     & $F_m$  & $Acc$    & $F_m$  & $Acc$   \\  \midrule
			(1) & $L_{seg}$      & 70.25\%  & 88.84\% &    50.36\% & 83.19\%     & 49.59\%  & 57.90\%  & 54.12\%  & 54.89\%  \\
			(2) & $L_{seg}+L_s+L_{conf}$           & 75.00\%  & 90.80\% &    58.27\% & 88.77\%     & 55.68\%  & 62.70\%  & 66.84\%  & 61.38\%   \\
			(3) & $L_{seg}+L_{c}+L_s+L_{conf}+L{con}$         & 76.04\%  & 91.23\% &    61.16\% & 88.68\%     & 70.55\%  & 79.91\%  & 68.19\%  & 64.50\%    \\
			(4) & $All$ $Modules$    & \textbf{80.70\%}  & \textbf{92.53\%} &   \textbf{68.05\%} & \textbf{90.00\%}     & \textbf{88.43\%}  & \textbf{88.22\%}  & \textbf{91.45\%}  & \textbf{91.21\%} 
			\\ \bottomrule 
		\end{tabular}
		\label{ab-loss}
	\end{table*}
	
	\begin{table}[!t]
		\centering
		\caption{\sc The Impact of Number of Stages in Our Method with Class-average F-score}
		\begin{tabular}{c|cccc}
			\toprule
			\textbf{Stages} & \textbf{Opportunity} & \textbf{PAMAP2} & \textbf{Hospital} & \textbf{Skoda}  \\\midrule
			$S=1$ &56.92\%  &78.67\%  &72.76\% &90.81\% \\
			$S=2$ &68.04\%  &\textbf{88.43\%}  &78.15\% &\textbf{91.45\%} \\
			$S=3$ &\textbf{68.05\%} &85.27\% &78.62\% &90.60\% \\
			$S=4$ &66.29\% &80.08\%  &\textbf{80.70\%}  &89.78\%  \\
			$S=5$ &66.14\% &79.59\%  &77.52\%  &89.13\%  \\
			\bottomrule 
		\end{tabular}
		\label{stages}
	\end{table}
	
	To comprehensively evaluate the efficacy of our proposed method, we conduct experiments on four challenging public Human Activity Recognition (HAR) datasets: Hospital \cite{yao2018efficient}, Opportunity \cite{chavarriaga2013opportunity}, PAMAP2 \cite{reiss2012introducing}, and Skoda \cite{stiefmeier2008wearable} datasets. For the weakly supervised setting with timestamp annotations, this paper employs timestamp labeling for training. Given that each sample in the current dataset is already labeled, it is essential to adapt this existing dataset to acquire timestamp labels. Our experiment first identifies all active segments within the sequence, then randomly selects a sample from each segment for annotation. To ensure the consistency and reproducibility of training results, we will subsequently disclose the timestamp annotations and release our codes for readers' convenience.
	\begin{itemize}
		\item \textbf{Hospital Dataset} \cite{yao2018efficient} is dedicated to capturing the continuous activities of elderly patients in a hospital environment. This dataset records natural and rich transitions in activities at a frequency of 10 Hz using a single IMU sensor. It encompasses 7 daily activities collected from 12 older patients admitted to the hospital. For our experiments, we designate Participants 1-8 for training, Participants 9-11 for testing, and participant 12 for validation.
		
		\item \textbf{Opportunity Dataset} \cite{chavarriaga2013opportunity} comprises measurements of acceleration, angular velocity, and magnetic force. This dataset defines 17 sporadic gestures along with a Null class, gathered from 4 participants at a frequency of 30 Hz. We select ADL-4 and 5 from Participants 2 and 3 for testing, ADL-2 from Participant 1 for validation, and the remaining data for training.
		
		\item \textbf{PAMAP2 Dataset} \cite{reiss2012introducing} provides sensory data from 9 participants engaged in 12 diverse activities, recorded at a frequency of 100 Hz. In our experiments, we use runs 1 and 2 from Participant 5 for validation, runs 1 and 2 from Participant 6 for testing, and the remaining data for training.
		
		\item \textbf{Skoda Dataset} \cite{stiefmeier2008wearable} targets a manufacturing scenario and includes 10 manipulative gestures. For our hold-out evaluation, we split the dataset into training, validation, and testing sets at a ratio of 8:1:1, following \cite{abedin2021attend}.
	\end{itemize}
	
	\subsection{Ablation Study}
	This section quantitatively verifies the effectiveness of different modules in our model. We conduct the ablation study from the composition of loss terms, the selection of hyper-parameters and the stage number of MS-TCN, and the generation of sample-level pseudo-labels.
	\begin{table}[!t]
		\centering
		\caption{\sc The Impact of Pseudo Label Generation}
		\begin{tabular}{c|cccc}
			\toprule
			& \textbf{Opportunity} & \textbf{PAMAP2} & \textbf{Hospital} & \textbf{Skoda}  \\\midrule
			Baseline + ABE &63.73\% &84.34\% &75.99\% &80.74\% \\
                Baseline + GCN &63.82\%  &82.52\%  &75.44\% &77.39\% \\
			Our Variants (1)  &63.56\%  &83.48\%  &76.33\% &90.08\% \\
			Our Variants (2)  &64.35\%  &80.93\%  &78.64\% &90.69\% \\
			Our method 	&\textbf{68.05\%} &\textbf{88.43\%}  &\textbf{80.70\%}  &\textbf{91.45\%}  \\
			\bottomrule 
		\end{tabular}
		\label{pseudo-gen}
	\end{table}
	\subsubsection{Impact of Loss Formulation}
	First, we investigate the effectiveness of each component in the loss composition of our model, as shown in Table \ref{ab-loss}. Referring to Section \ref{method}, we compare four variations: (1) the base model is trained solely using timestamp labels and the sample-level classification loss (Eq.\eqref{lseg}); (2) the primary network is trained to incorporate smooth loss and confidence loss (Eq.\eqref{eqxx1}); (3) the model is trained with the addition of multi-label soft margin loss and sample-to-prototype contrastive loss (Eq.\eqref{final_sample}); (4) our fully proposed method, complete with all modules (Eq.\eqref{final_loss}).
	
	From Table\ref{ab-loss}, we could find that these modules have a positive effect on performance improvement. With the results of $Acc$ and $F_m$, our fully proposed method (4) performs best on all four evaluation datasets. By comparing the performance of (2) and (3), these two additional losses (multi-label soft margin loss $L_{cls}$ and sample-to-prototype contrastive loss $L_{con}$) contribute to improved performance, especially on the PAMAP2 dataset, where the $F_m$ has increased by 14.87\%. Moreover, an analysis of the performance gains between (3) and (4) reveals that using optimal transport theory for pseudo-label generation significantly enhances recognition and segmentation performance, particularly noticeable over the PAMAP2 and Skoda datasets, which have long durations of single activities. As only a single sample is annotated within an activity segment, datasets with fewer activity transitions (or those where each activity lasts longer) will have a lower proportion of annotated samples. Furthermore, we analyze the impact of different pseudo-label generation methods on recognition and segmentation performance in Section \ref{PLG}. The smooth loss $L_s$ and confidence loss $L_{conf}$ have been widely proven effective in various video- and sensor-based works \cite{li2021temporal, 9772403}. In our experiment, the smoothness and confidence loss would also bring significant improvements, which are increased by 4.75\%/7.91\%/6.09\%/12.72\% on the Hospital/Opportunity/PAMAP2/Skoda datasets, respectively.
	
	\begin{figure}[t]
		\centering
		\subfigure[The optimal $\lambda_{con}$ on Skoda dataset]{\includegraphics[width=2.9cm]{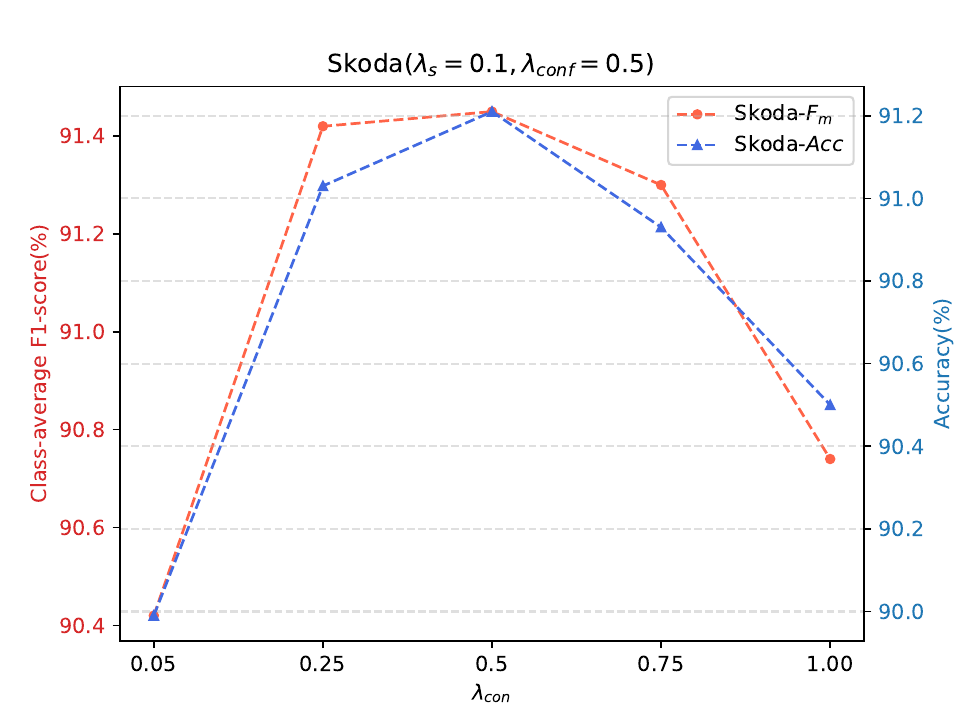}}
		\label{S-con}
		\subfigure[The optimal $\lambda_{s}$ on Skoda dataset]{\includegraphics[width=2.9cm]{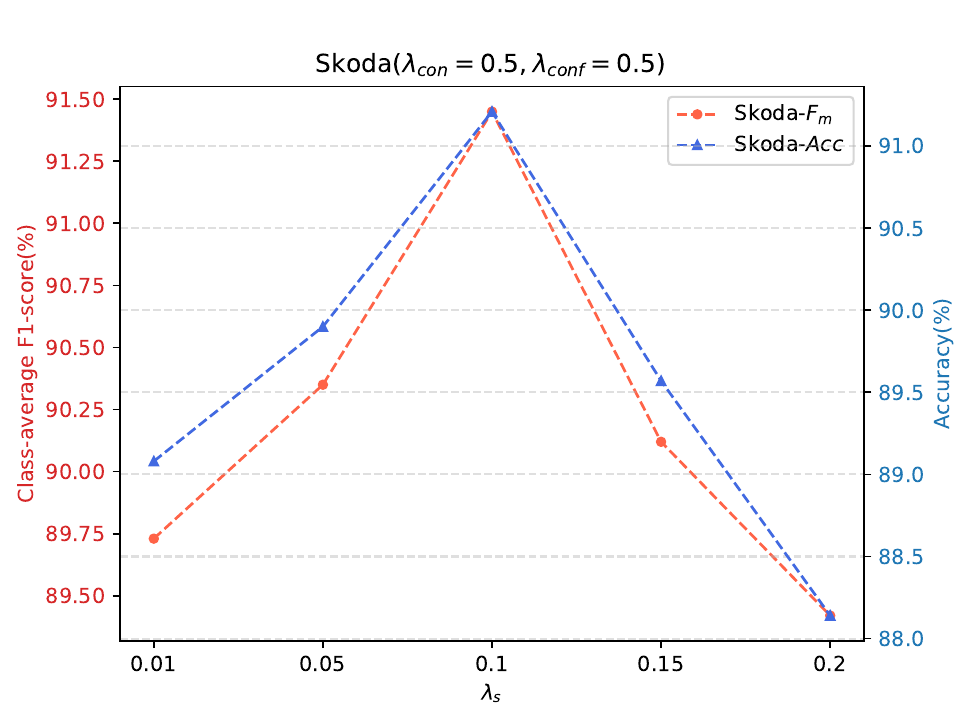}}
		\label{S-s}
		\subfigure[The optimal $\lambda_{conf}$ on Skoda dataset]{\includegraphics[width=2.9cm]{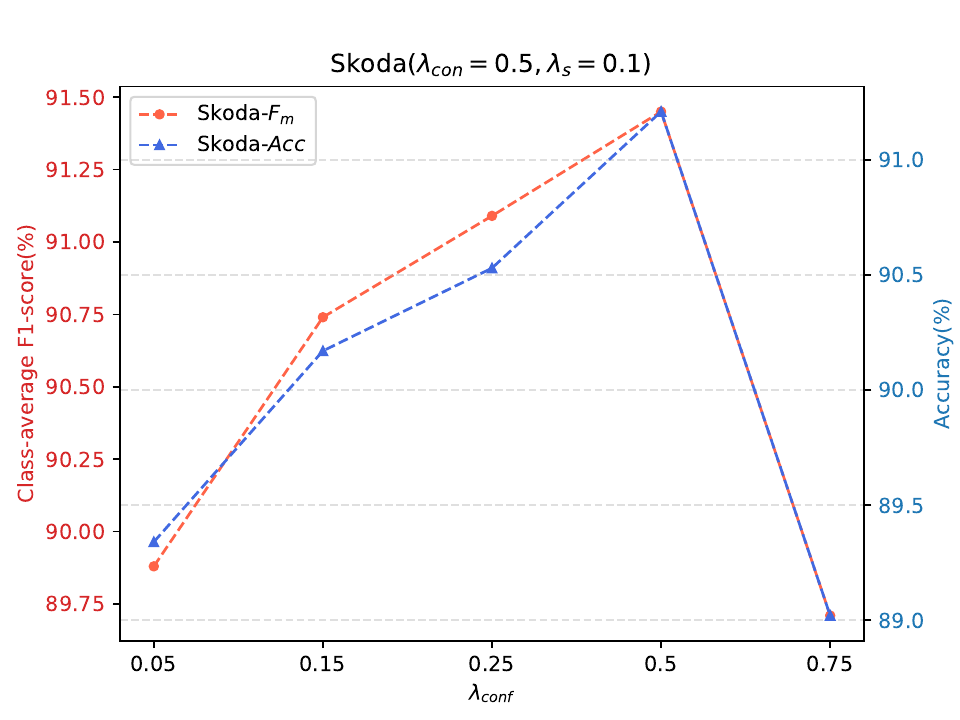}}
		\label{S-conf}\\
		\subfigure[The optimal $\lambda_{con}$ on Hospital dataset]{\includegraphics[width=2.9cm]{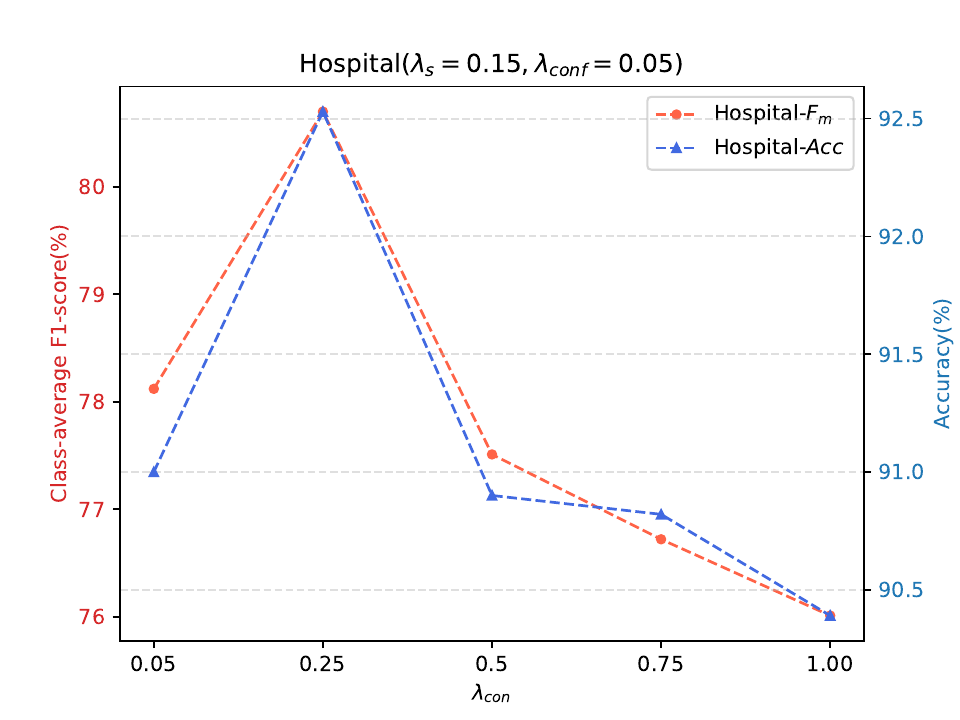}}
		\label{H-con}
		\subfigure[The optimal $\lambda_{s}$ on Hospital dataset]{\includegraphics[width=2.9cm]{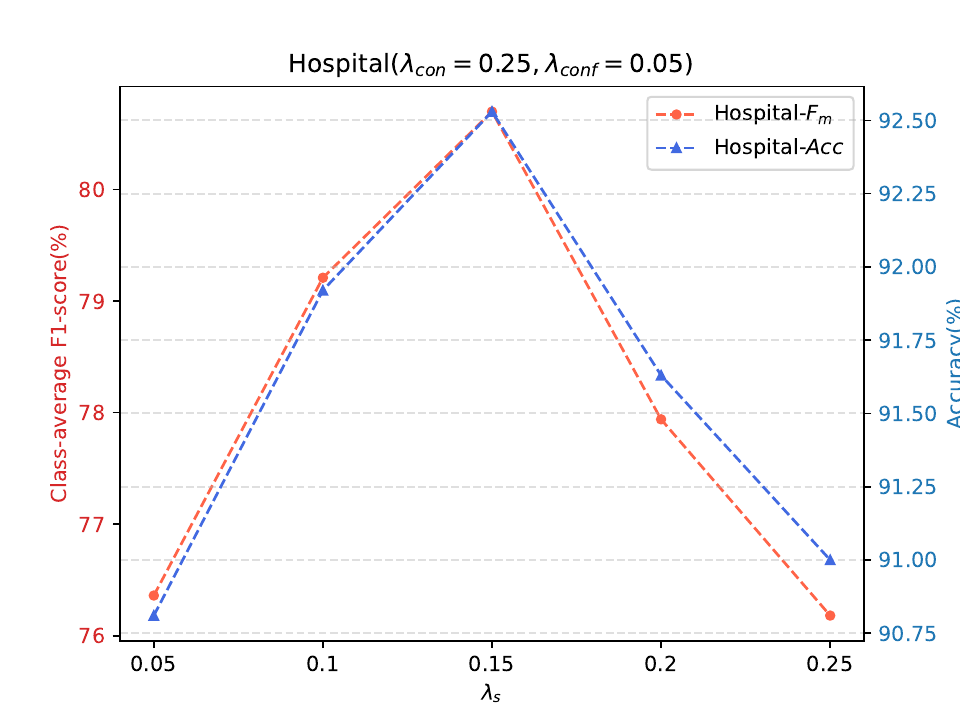}}
		\label{H-s}
		\subfigure[The optimal $\lambda_{conf}$ on Hospital dataset]{\includegraphics[width=2.9cm]{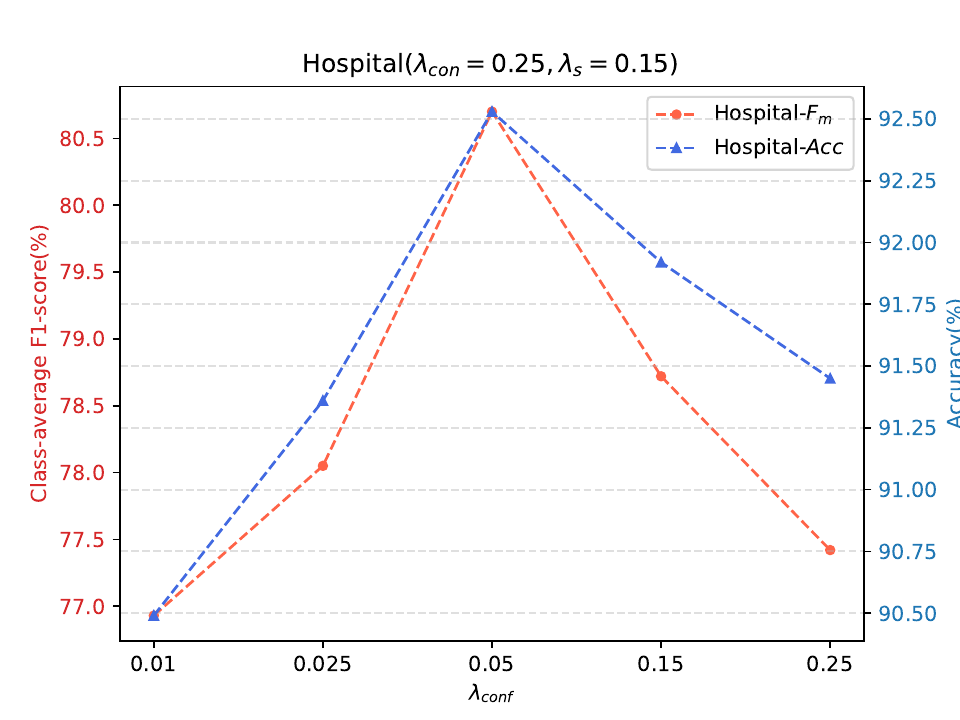}}
		\label{H-conf}
		\caption{The effects on the three hyper-parameters $\lambda_{con}$, $\lambda_{s}$, $\lambda_{conf}$ on Skoda and Hospital dataset. }
		\label{hyper-parameters}
	\end{figure}

	\subsubsection{Impact of Hyper-parameters}
	To balance the contribution of different modules for better segmentation and recognition performance, our proposed model sets three hyper-parameters $\lambda_{con}$, $\lambda_{s}$, $\lambda_{conf}$, as shown in Eq.\eqref{final_loss}. We expect to find the optimal performance by varying the values of these hyper-parameters while other hyper-parameters are fixed, as shown in Fig. \ref{hyper-parameters}. We choose the Skoda and Hospital dataset for experiments and use the class-average F-score and Accuracy to evaluate the recognition performance for obtaining the optimal hyper-parameters.
	
	As shown in Fig.\ref{hyper-parameters}, we can derive the optimal parameters for our method on the Skoda and Hospital datasets by individually adjusting each of the three parameters. On the Skoda dataset, we achieve the best result at $\lambda_{con}=0.5$, $\lambda_{s}=0.1$, $\lambda_{conf}=0.5$, while on the Hospital dataset, the optimal performance is achieved at $\lambda_{con}=0.25$, $\lambda_{s}=0.15$, $\lambda_{conf}=0.05$. Although the optimal results can be found on both Skoda and Hospital datasets by adjusting hyper-parameters, the choice of hyper-parameters is greatly affected by the dataset. Hence, through additional experiments, we discerned the optimal hyperparameters for the other two datasets used in our study: Opportunity and PAMAP2. For the PAMAP2 dataset, we set the hyperparameters at $\lambda_{con}=0.5$, $\lambda_{s}=0.25$, and $\lambda_{conf}=0.75$. For the Opportunity dataset, we selected $\lambda_{con}=0.75$, $\lambda_{s}=0.15$, and $\lambda_{conf}=0.75$ as the optimal hyperparameters.
	
	\subsubsection{Impact of Multiple Stages}
	In line with our preceding work \cite{9772403}, our methodology also relies on the Multi-Stage Temporal Convolutional Network (MS-TCN) to create a multi-stage architecture. Selecting an appropriate number of stages can significantly enhance the model's performance. In this experiment, we explore the performance across stages 1 to 5 on the four public wearable-based HAR datasets with a weakly supervised setting, as demonstrated in Table \ref{stages}.
	
	In Table \ref{stages}, we can find the optimal number of stages for our method on these four publicly available datasets, where the number of stages that achieve optimal results varies across datasets. For the PAMAP2 and Skoda datasets, the optimal stage number is 2, yielding results of 88.34\% and 91.45\% on class-average F-score. On the Opportunity and Hospital datasets, the optimal number of stages is 3 and 4, respectively. This variability underscores the importance of customizing the number of stages to the specific characteristics of each dataset.
	
	\subsubsection{Impact of Pseudo Label Generation}
	\label{PLG}
	In this experiment, we comprehensively compare our proposed pseudo-label generation module with existing methods and additionally compare various variants of our pseudo-label generation method. For comparative purposes, we have selected two existing pseudo-label generation methods: (1) Generation of sample-level pseudo-labels using a forward-backward action change detection module by minimizing the stamp-to-stamp energy function\cite{li2021temporal} (Baseline+ABE); (2) Generation of sample-level pseudo-labels utilizing graph convolutional networks\cite{khan2022timestamp} (Baseline+GNC). We could get performance on these four datasets as shown in Table \ref{pseudo-gen}.
	
	Furthermore, we consider two variants of our proposed method for comparison: Variant I: the pseudo-hard labels are directly obtained according to the estimated optimal transport matrix, regardless of the order of labels in the activity segment; Variant II: the pseudo-labels are assigned based on the proportion of numbers belonging to the two classes of the activity segment in the estimated optimal transport matrix. In Table \ref{pseudo-gen}, we employ the class-average F-score to illustrate the performance of these methods. Our method surpasses the comparison algorithms, demonstrating the effectiveness of our pseudo-label generation approach. Moreover, compared to these two variants, the hybrid pseudo-label generation strategy of hard and soft labels enhances activity recognition performance. This design signifies that our model's adeptness in handling pseudo-labels offers a substantial contribution towards improved activity recognition results.
	
	\begin{table*}[!t]
		\centering
		\caption{\sc The Overall Performance on the Four Datasets with Fully and Weakly Supervised Setting}
		\begin{tabular}{c|c|cc|cc|cc|cc}
			\toprule
			\multicolumn{1}{c|}{\multirow{2}{*}{\textbf{Supervision}}} &\multicolumn{1}{c|}{\multirow{2}{*}{\textbf{Competing Method}}} & \multicolumn{2}{c|}{\bfseries{Hospital}} & \multicolumn{2}{c|}{\bfseries{Opportunity}} & \multicolumn{2}{c|}{\bfseries{PAMAP2}}  & \multicolumn{2}{c}{\bfseries{Skoda}}\\ \cmidrule{3-4}  \cmidrule{5-6}  \cmidrule{7-8} \cmidrule{9-10} 
			& & $F_m$  & $Acc$      & $F_m$  & $Acc$    & $F_m$  & $Acc$  & $F_m$  & $Acc$  \\  \midrule
			\multicolumn{1}{c|}{\multirow{4}{*}{Fully}} & Attn.\cite{murahari2018attention} &  64.37\% &  89.01\%  & 70.85\% & 91.70\% & 87.41\% & 90.32\%  & 91.42\% &91.29\%\\
			& A\&D \cite{abedin2021attend} & 66.63\% & 89.52\%  &74.61\%   & 92.65\%   &  90.83\%   & 90.55\%   & 92.78\% & 92.62\% \\
			& JSR-TII\cite{9772403} & \underline{81.64\%} & \underline{93.02\%} & \textbf{78.69\%} & \textbf{93.66\%}  & \underline{93.26\%} & \underline{93.81\%}  & \underline{94.74\%} & \underline{95.01\%} \\
			& Our-Full-supervised & \textbf{82.04}\% & \textbf{93.94}\% & \underline{77.19}\% & \underline{93.03}\% &  \textbf{93.77\%}    &\textbf{94.15\%}   & \textbf{95.06}\% & \textbf{95.29}\% \\ \midrule
			\multicolumn{1}{c|}{\multirow{4}{*}{Weakly}} & TAS-ABE\cite{li2021temporal} &\underline{75.42\%}  &90.09\%  &64.14\%  &87.96\%  &77.44\% &\underline{85.30\%} &\underline{81.73\%} &78.79\% \\
			&TAS-MLP \cite{khan2022timestamp} &73.75\% &\underline{90.42\%} &60.89\% &\underline{88.40}\% &75.15\% &76.99\% &72.29\% &66.88\% \\
			&TAS-GCN \cite{khan2022timestamp} &74.28\% &88.21\% &\underline{65.62\%} &87.97\% &\underline{81.45\%} &82.94\% &78.56\% &\underline{81.28\%} \\
			&Our method   &\textbf{80.70\%} & \textbf{92.53\%}  &\textbf{68.05\%} & \textbf{90.00\%}	&\textbf{88.43\%} &\textbf{88.22\%}  &\textbf{91.45\%}  &\textbf{91.21\%}  \\
			\bottomrule 
		\end{tabular}
		\label{SOTA}
	\end{table*}
	
	\begin{table*}[!t]
		\centering
		\caption{\sc The Segmentation Results on the Four Public Datasets}
		\begin{tabular}{c|ccc|ccc|ccc|ccc}
			\toprule
			\multicolumn{1}{c|}{\multirow{2}{*}{\textbf{Methods}}} & \multicolumn{3}{c|}{\bfseries{Hospital}} & \multicolumn{3}{c|}{\bfseries{Opportunity}} & \multicolumn{3}{c|}{\bfseries{PAMAP2}}  & \multicolumn{3}{c}{\bfseries{Skoda}}\\ \cmidrule{2-4}  \cmidrule{5-7}  \cmidrule{8-10} \cmidrule{11-13} 
			& $JI \uparrow$  & $IoU \uparrow$ & $O/U \downarrow$  & $JI \uparrow$  & $IoU \uparrow$ & $O/U \downarrow$ & $JI \uparrow$  & $IoU \uparrow$ & $O/U \downarrow$  & $JI \uparrow$  & $IoU \uparrow$ & $O/U \downarrow$ \\  \midrule
			
			TAS-ABE \cite{li2021temporal}&62.97\% &59.14\% &7.15\% &48.40\% &54.70\% &6.83\% &69.29\% &67.76\% &4.11\% &70.45\% &56.78\% &19.50\%\\
			TAS-GCN \cite{khan2022timestamp} &62.18\%  &58.30\%  &7.57\%  &50.85\%  &59.41\% &6.05\% &73.31\% &63.14\% &5.86\% &69.15\% &56.76\% &12.59\% \\
			Ours &\underline{69.50\%} &\underline{63.24\%} &\underline{5.30\%} &\underline{53.13\%} &\underline{60.55\%} &\underline{5.52\%} &\underline{80.73\%} &\underline{70.31\%} &\underline{3.74\%} &\underline{84.93\%} &\underline{75.99\%} &\underline{5.86\%}\\
			Ours-fully   &\textbf{72.58\%} & \textbf{65.83\%}  &\textbf{4.70\%} & \textbf{63.89\%}	&\textbf{67.41\%} &\textbf{3.73\%}  &\textbf{88.64\%}  &\textbf{84.54\%} &\textbf{1.34\%} &\textbf{91.01\%}  &\textbf{85.41\%}  &\textbf{2.99\%} \\
			\bottomrule 
		\end{tabular}
		\label{SOTA-Seg}
	\end{table*}
\begin{figure*}[ht]
	\centering
	\subfigure[ABE]{\includegraphics[width=5.9cm]{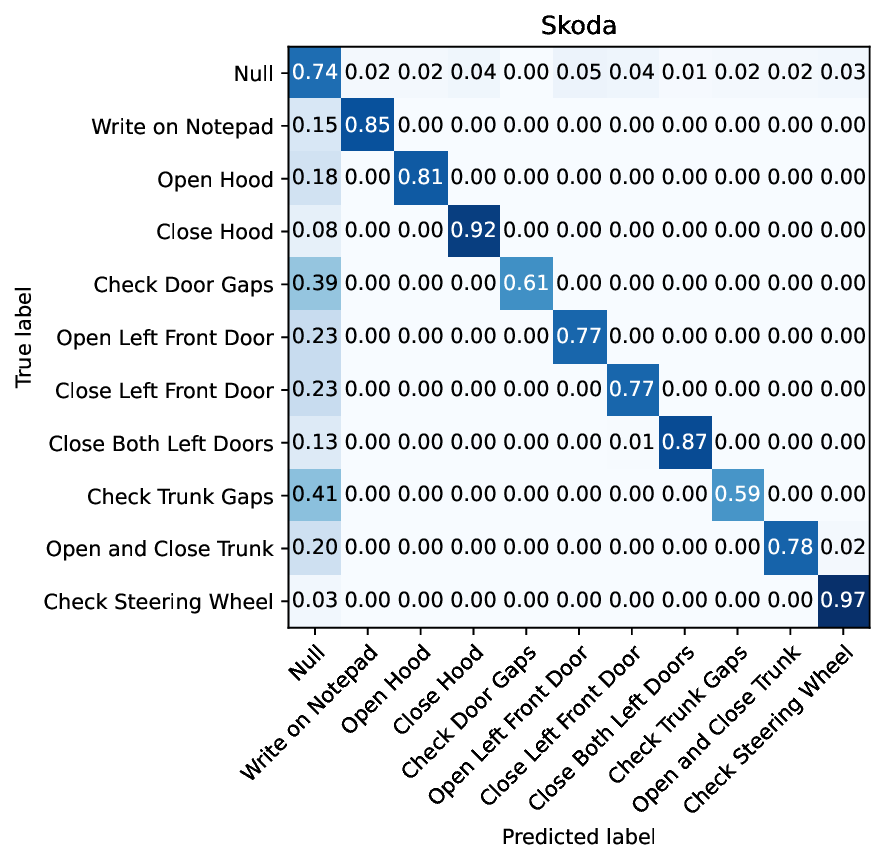}}
	\label{M-ABE}
	\subfigure[Ours]{\includegraphics[width=5.9cm]{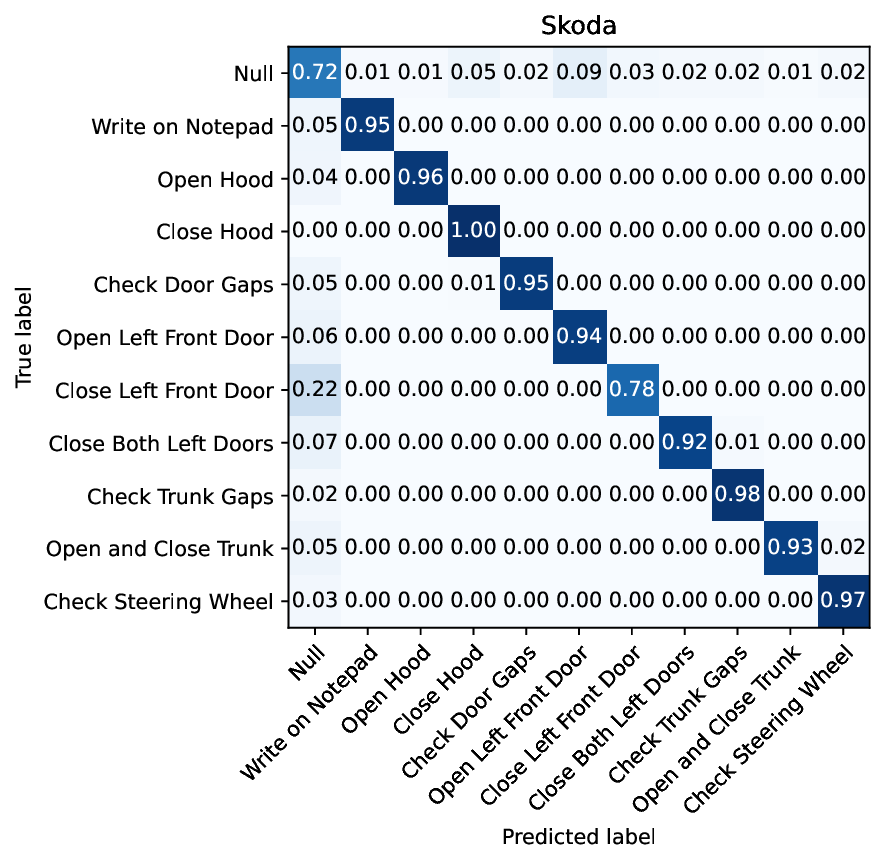}}
	\label{M-ours}
	\subfigure[Ours-fully]{\includegraphics[width=5.9cm]{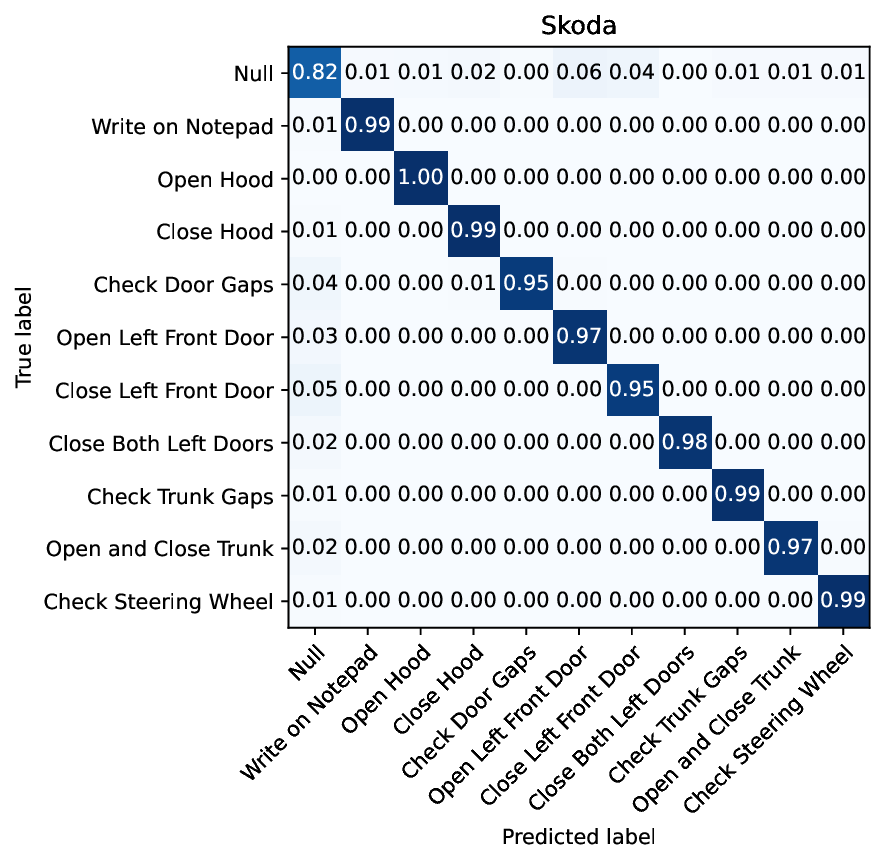}}
	\label{M_ours-fully}
	\caption{The confusion matrices of different methods on the Skoda dataset. (a) The ABE model (weakly supervised setting). (b) The proposed method (weakly supervised setting). (c) The proposed method (in a typical supervised setting).}
	\label{Confusion-matrix}
\end{figure*}
	\begin{figure*}[ht] 
		\subfigure[Qualitative results on the Hospital (Samples 200 -2000).]{
			\includegraphics[width=0.49\textwidth]{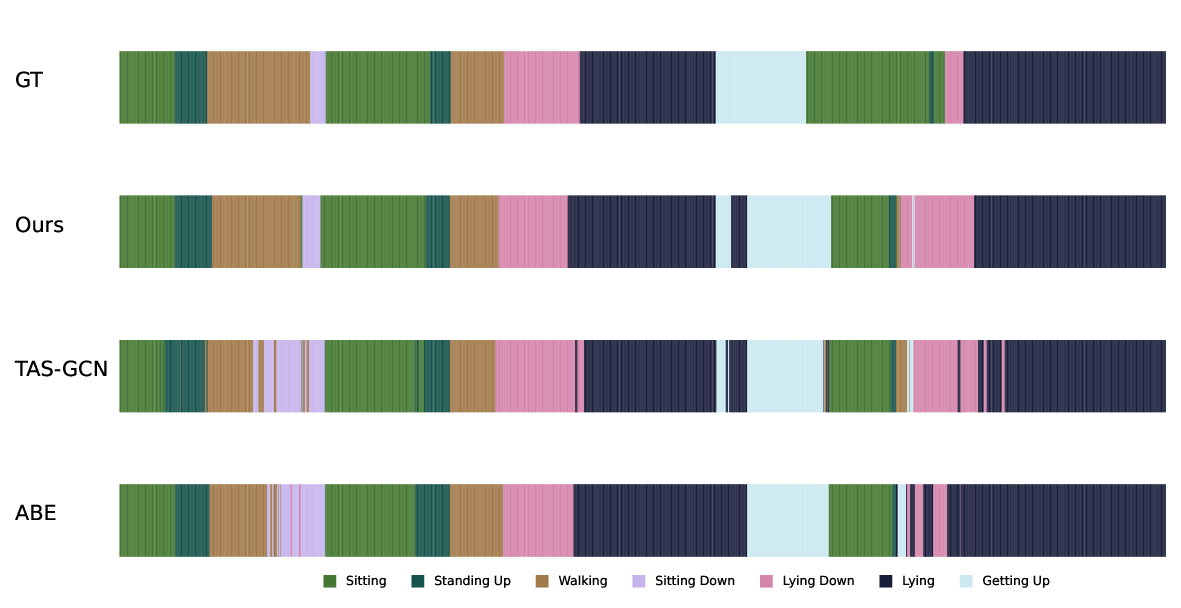}} 
		\subfigure[Qualitative results on the Skoda (Samples 1500-3500).]{
			\includegraphics[width=0.49\textwidth]{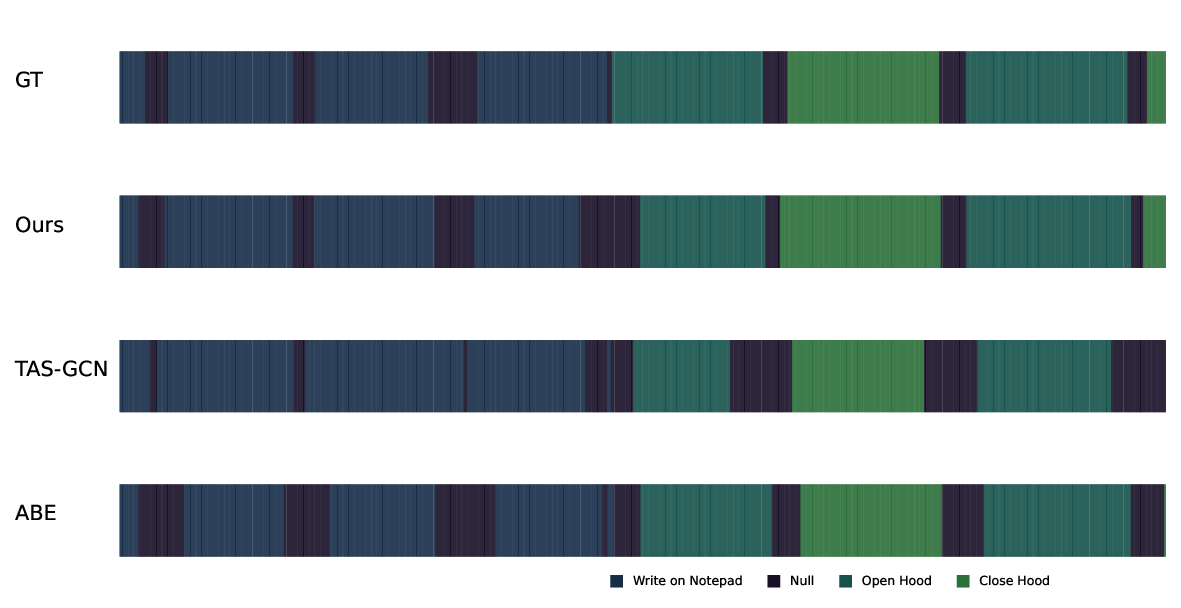}} 
		
		\subfigure[Qualitative results on the Opportunity (Samples 9000 - 14000).]{
			\includegraphics[width=0.49\textwidth]{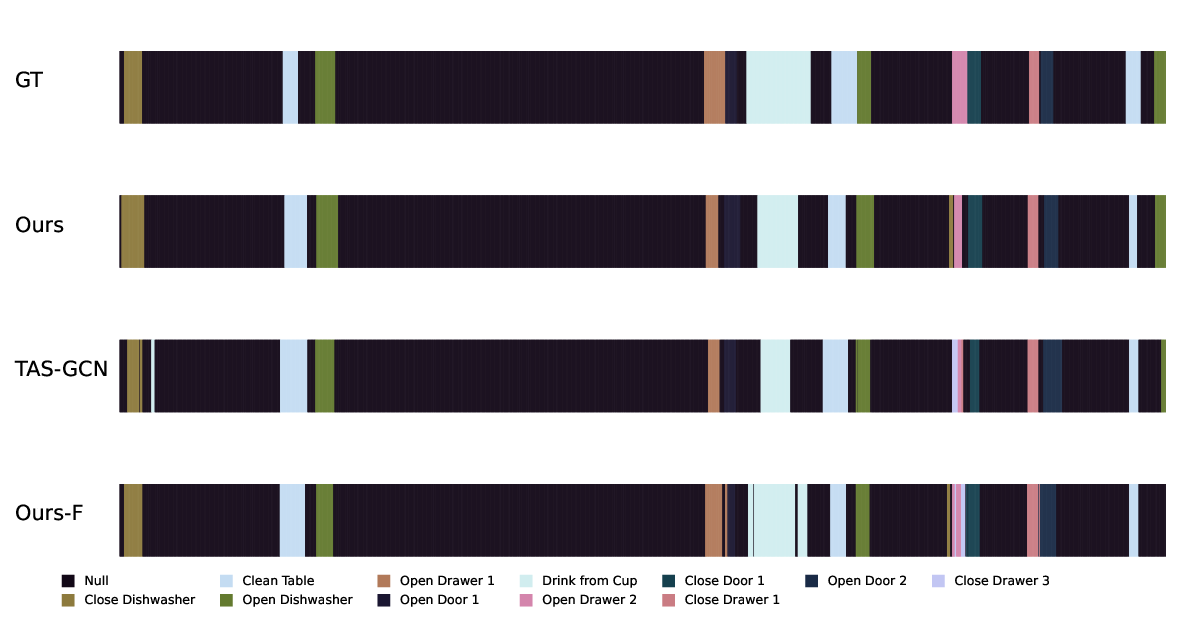}} 
		\subfigure[Qualitative results on the PAMAP2 (Samples 30 - 16000).]{
			\includegraphics[width=0.49\textwidth]{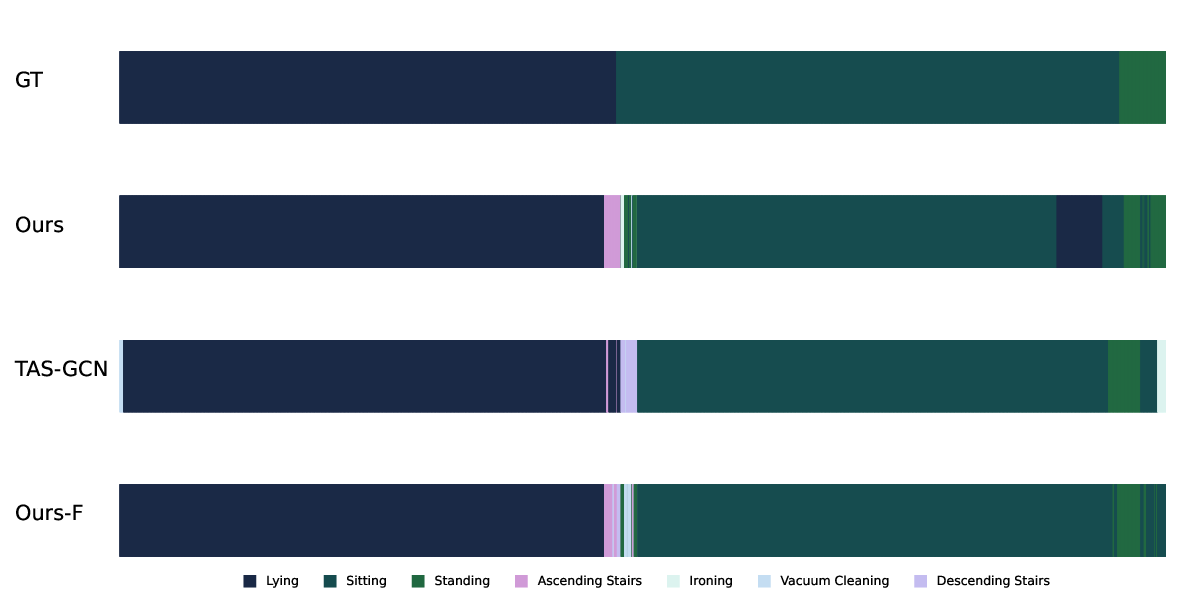}} 
		
		\caption{Qualitative results visualization on the Hospital and Opportunity datasets. (a), (b) in comparison with the state-of-the-art competing methods. (c) and (d) are the success and failure cases compared to sample-level predictions before refinement.}
		\label{DIngxing}
	\end{figure*}
	
	\subsection{Comparison with the State-of-the-Art}
	\label{CSOTA}
	In Section \ref{CSOTA}, we compare the segmentation and recognition performance of the proposed model with the state-of-the-art competing methods on the four public wearable-based HAR datasets. Under the fully supervised and weakly supervised setting, some recently advanced algorithms are selected for overall comprehensive comparison, which are listed as follows:
	
	1) Fully supervised methods: Current state-of-the-art activity recognition or joint segmentation and recognition methods with a fully supervised setting.
	\begin{itemize}
		\item \textbf{JSR} \cite{9772403}: A joint activity segmentation and recognition model with wearable sensors, which forms a multi-task learning framework with a boundary prediction and boundary consistency module.
		\item \textbf{A\&D} \cite{abedin2021attend}: A novel deep learning method for HAR, which contains the CIE, AGE, and center-loss module. And it has obtained the SOTA recognition performance through a data enhancement approach.
		\item \textbf{Attn.} \cite{murahari2018attention}: An effective deep learning HAR model, which incorporates attention layers into the DeepConvLSTM.
	\end{itemize}
	
	2) Weakly supervised methods: The state-of-the-art weakly supervised activity segmentation methods with timestamp labels. 
	\begin{itemize}
		\item \textbf{TAS-ABE} \cite{li2021temporal}: A temporal action segmentation model for video sequence from timestamp supervision, which generates the frame-level pseudo-labels by detecting the action changes.
		
		\item \textbf{TAS-GCN} \cite{khan2022timestamp}: A novel method learns a graph convolutional network to exploit both frame features and connections between neighboring frames to transform sparse timestamp labels to dense frame-level pseudo-labels.
		
		\item \textbf{TAS-MLP} \cite{khan2022timestamp}: A novel method learns the MLP network to generate dense frame-level pseudo-labels from sparse timestamp labels.
	\end{itemize}
	
	Table \ref{SOTA} provides a comprehensive comparison of our proposed method with other techniques under both fully and weakly supervised settings. For a fair comparison, we use the sample-wise evaluation protocol \cite{abedin2021attend, 9772403} for all comparing methods. Despite the fact that our method is primarily designed for a weakly supervised setting, it remains competitive under a fully supervised setting. On the Hospital, Skoda, and PAMAP2 datasets, our approach achieves the best recognition performance, recording 82.05\%, 95.06\%, and 93.77\% class-average F-scores, respectively. For the weakly supervised setting with only timestamp labels, our approach surpasses all state-of-the-art competitors, improving the performance by 5.28\%, 2.43\%, 6.98\%, and 9.72\% on the Hospital, Opportunity, PAMAP2, and Skoda datasets with class-average F-scores, respectively. 
	
	Furthermore, we employ segmentation metrics ($JI$, $IoU$, $O/U$) to comprehensively illustrate our method's segmentation performance, which provides a more accurate reflection of performance in natural, unconstrained environments, as displayed in Table \ref{SOTA-Seg}. Our method achieves state-of-the-art segmentation performance under full supervision, reaching the 72.58\%/63.89\%/88.64\%/91.01\% with $Jaccard$ $Index$ on the Hospital/Opportunity/PAMAP2/Skoda datasets, respectively. Moreover, in a weakly supervised setting, our method outperforms the other weakly supervised methods, outperforming the sub-optimal results by 6.53\%/2.28\%/7.42\%/14.48\% on these four datasets. For the metrics $O/U$, our method also achieves state-of-the-art results on all datasets, demonstrating that the recognition and segmentation performance of our method on the activity transition boundary remains stable. Especially in the Hospital dataset with frequent transitions of activities, ours with weakly supervised setting improves by 6.53\%/4.1\%/1.85\% respectively compared to the sub-optimal method (TAS-ABE) in terms of $JI$, $IoU$ and $O/U$. Next, we will further demonstrate the segmentation performance using the quantitative analysis results in Section \ref{qualitative}.
	
	\subsection{Qualitative Evaluation}
	\label{qualitative}
	To more fully demonstrate the superiority of our proposed method, we illustrate through the following four qualitative experiments: (1) Confusion matrix; (2) Segmentation qualitative results; (3) Feature embedding visualization with t-SNE; (4) Class Activate Maps (CAMs) visualization. 
	
	\subsubsection{Analysis on the Confusion Matrix}
	In Fig.\ref{Confusion-matrix}, we show the confusion matrices of our proposed method and competing method on the Skoda dataset. The confusion matrices of the $ABE$ method and our proposed approach with a weakly supervised setting are shown in Fig.\ref{Confusion-matrix} (a) and (b), respectively. The confusion matrix allows us to discern the degree of accuracy our approach achieves for most activity categories. It is noticeable that the dominated activity categories identified by the competing method exhibit a significant degree of confusion with the 'Null' category. However, our method is more adept at avoiding these errors. Notably, our approach demonstrates considerable performance enhancements in the 'Check Door Gaps' and 'Check Trunk Gaps' categories compared to the competing $ABE$ algorithm, where the accuracy in these categories has increased by 34\% and 39\%, respectively.
	
	Furthermore, we show the confusion matrix of our approach trained with the fully supervised setting on the Skoda dataset. Comparison between Fig.\ref{Confusion-matrix} (b) and (c) reveals an overall improvement in the recognition performance across most activity categories. Significantly, the class 'Close Left Front Door' improved by 17\% compared to our model trained with timestamp labels. Conversely, our model can also perform relatively well in a weakly supervised setting.
	
	\subsubsection{Segmentation Qualitative Results}
	In this section, the sample-level activity predictions are visualized in Fig. \ref{DIngxing}, where different colors represent different activity categories. We have selected several sequence fragments from the four public HAR datasets, and the quantitative results of our method, along with those of TAS-ABE \cite{li2021temporal} and TAS-GCN \cite{khan2022timestamp}, are presented for comparison. As illustrated in Fig.\ref{DIngxing} (a) and (b), we present the sample-level predictions of our method in comparison with competing methods with a weakly supervised setting. Our method effectively reduces over-segmentation errors, which refer to the fluctuating inaccuracies in sample-level predictions within each activity segment, where the improvement over competing methods is clearly demonstrated in Fig.\ref{DIngxing} (a). For the fragments in the Skoda dataset in Fig.\ref{DIngxing} (b), we find that our method could get the sample-level activity predictions more accurately on most activity transition boundaries, which corresponds to our method achieving state-of-the-art segmentation performance on the $O/U$ metric in Table\ref{SOTA-Seg}. The last row of Fig.\ref{DIngxing} (c) and (d) show the sample-level activity predictions obtained by our proposed model trained with the fully supervised setting. As can be observed from these visualizations, employing all sample-level activity labels (fully supervised setting) would improve both activity segmentation and recognition. Notably, our model trained using only timestamp labels delivers promising results that closely approach those achieved with the fully supervised setting. These visualizations demonstrate the effectiveness of our method even in a weakly supervised setting.
	
	\begin{figure}[t]
		\centering
		\subfigure[TAS-GCN]{\includegraphics[width=4.2cm]{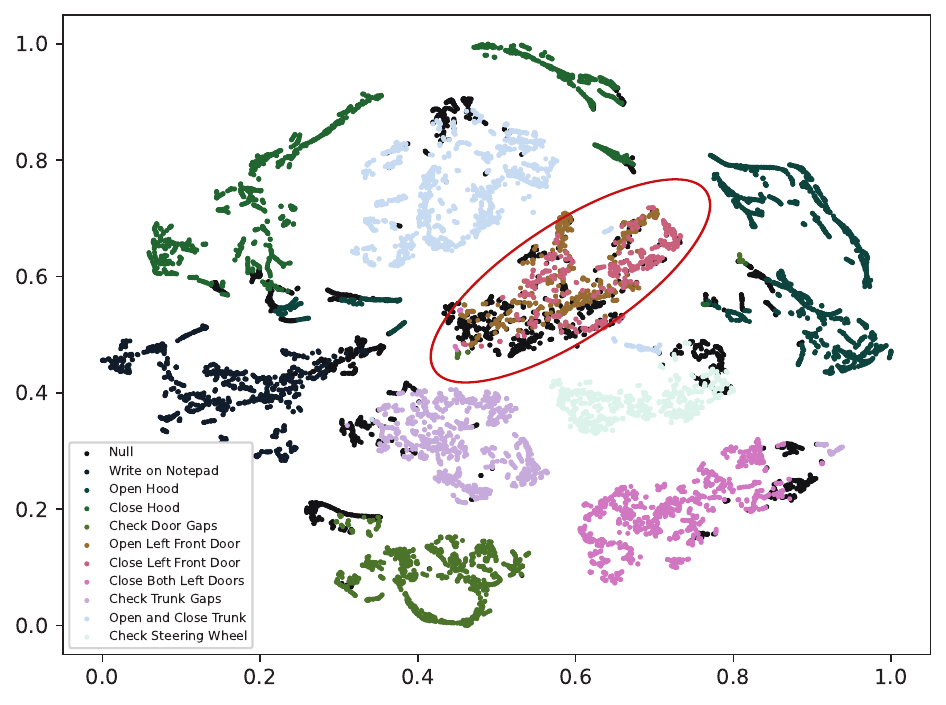}}
		\label{gcn-skoda}
		\subfigure[Ours]{\includegraphics[width=4.2cm]{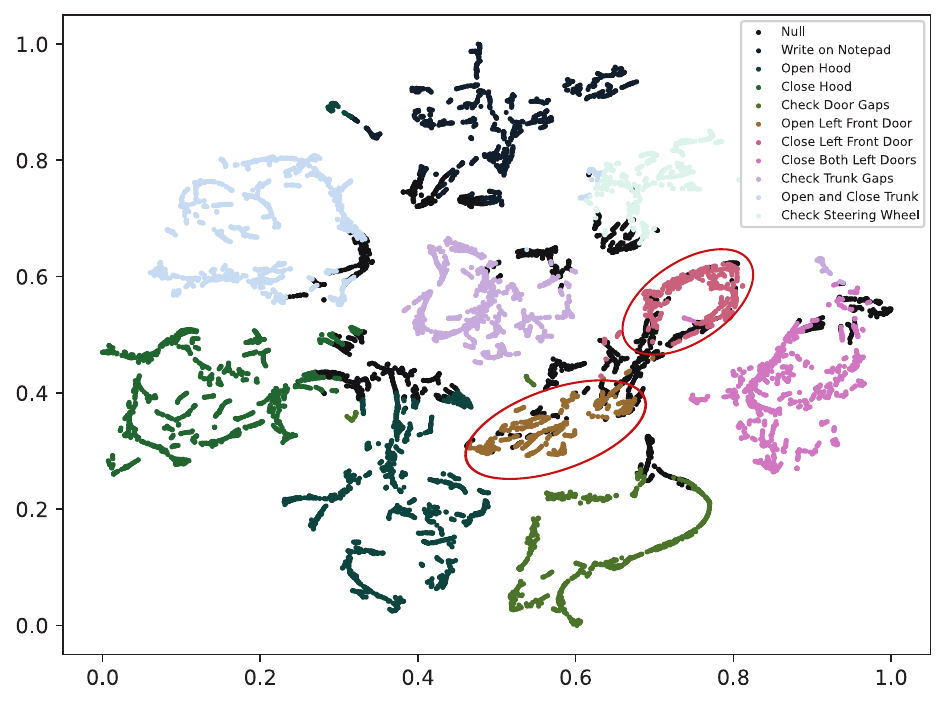}}
		\label{our-skoda}
		\caption{The t-SNE visualization of our method and TAS-GCN on the Skoda dataset. (a) The t-SNE visualization of the TAS-GCN learned representation. (b) The t-SNE visualization of our method learned representation.}
		\label{Skoda-TSNE}
	\end{figure}
	\subsubsection{Analysis on the Feature Embedding Visualization with t-SNE}
	In this section, the t-distributed stochastic neighbor embedding (t-SNE) visualizations of the latent representation extracted by different methods are shown in Fig.\ref{Skoda-TSNE}. Each point in the figure is color-coded according to its associated activity label, while the position of each point represents the distribution of the feature dimensionality reduction of the model in a 2-dimensional space. In Fig.\ref{Skoda-TSNE} (a) and (b), both methods successfully distinguish most activity categories to a certain extent. However, our method exhibits fewer ambiguities in several confusable activity categories. These confusing activities, namely, 'Null', 'Open Left Front Door', and 'Close Left Front Door', are circled in red within the figure. For these three activities, the latent representations derived by our method show superior separability in the t-SNE visualizations, which further underscores the superiority of our proposed method.
	
	\subsubsection{The CAMs Visualization Results}
	
	\begin{figure}[t]
		\centering
		\includegraphics[width=9cm]{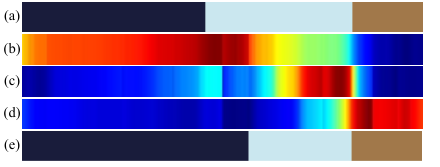}
		\caption{{The Class Activate Maps Visualization of the Fragment on the Hospital Dataset. (a) The ground truth for this fragment on the Hospital dataset. (b) The CAMs visualization for the 'Lying' class. (c) The CAMs visualization for the 'Getting Up' class. (d) The CAMs visualization for the 'Walking' class. (e) The qualitative predictions on the Hospital dataset. }}
		\label{CAMs}
	\end{figure}
	We further demonstrate the capability of our proposed method through Class Activation Maps (CAMs), as shown in Figure \ref{CAMs}. We selected a fragment from the Hospital dataset, which contains three distinct activities: 'Lying', 'Getting Up', and 'Walking'. The ground-truth labels for these activities and the predictions from our model are displayed in Figure \ref{CAMs} (a) and (e), respectively. Fig.\ref{CAMs} (b), (c), and (d) display the activation regions for the three activity categories within the selected fragment, where the intensity of the red color indicates the probability of the corresponding activity category within that region. As evident from the visualizations, the activation regions in the figures correspond significantly with the ground-truth labels within the fragment. However, we can notice the discrepancies and confusion around the boundaries between different activities. These visualizations of CAMs underline the efficiency of our model's multi-label classifier while also identifying areas for further enhancement, particularly for improving the accuracy of activity transition recognition.

	\subsection{Further Analysis}
	To further analyze the performance of our proposed method, we additionally discuss the computational complexity of various methods and the segmentation and  recognition performance in the mixed supervision setting.
	
	\subsubsection{Computational Complexity Analysis}
	We employ this experiment for computational complexity analysis on a PC platform with a GPU RTX 3090 and a CPU i9-13900K, where the batch size would be set as 32 and 1 for training and testing, respectively. For the traditional methods with a fixed-size sliding window, the window size and stride are set as 24/24 and 12/1 for the training/testing phase. The joint segmentation and recognition methods set the input length to 256.
	
	Table \ref{time} lists the computational complexity analysis results of different methods. As shown in the table, our method takes the shortest training and test time, and the number of our model parameters is also optimal. However, the weakly supervised comparison method $ABE$ \cite{li2021temporal} is time-consuming due to the method of estimating the activity boundary during the training phase. This further validates that the superior performance of our method is attributed to the design of the model and training strategy rather than the complexity of the model itself.
	\begin{table}[t]
		\centering
		\caption{\sc The Complexity Comparision of the Competing Methods on the PAMAP2 Dataset }
		\begin{tabular}{c|c|c|c}
			\toprule
			\textbf{Methods}         & \textbf{Param} & \makecell[c]{\textbf{Training}\\ \textbf{time (s/epoch)} } & \makecell[c]{\textbf{Inference}\\ \textbf{time (s/epoch)} } \\
			\midrule
			
			\textbf{Attn. \cite{murahari2018attention}}            & 1.983 M & 20.06         & 8.22                        \\
			\textbf{A\&D\cite{abedin2021attend}}          & 1,503 M & 25.61         & 11.88                      \\
			\textbf{JSR\cite{9772403}}             & 2.799 M & 12.11         & 0.78                       \\
			\textbf{TAS-GCN\cite{khan2022timestamp}}             & 0.590 M & 18.13         & 0.49                        \\
			\textbf{TAS-ABE\cite{li2021temporal}}             & 0.587 M & 100.03         & 0.11                      \\
			\textbf{Ours}             & \textbf{0.353 M} & \textbf{8.62}         & \textbf{0.08}                       \\
			
			\bottomrule             
		\end{tabular}
		\label{time}
	\end{table} 
	
	\subsubsection{Performance in Mixed Supervision Setting}
	Next, we delve into the performance of the proposed method under a mixed supervision setting, leveraging both timestamp labels and a certain fraction of sample-level labels. The outcomes associated with varying fractions of sample-level labels on the Skoda dataset are presented in Table \ref{mixed-supervision}. In the table, 0\% signifies that no additional training samples are used, where we solely rely on TimeStamp (\textbf{TS}) labels. 100\% corresponds to fully supervised learning. The remaining ratios (20\%-80\%) indicate an incremental involvement of a corresponding fraction of samples within each segment in the training process. As the fraction of sample-level labels involved increases, we observe an enhancement in our method's performance. Nevertheless, when the fraction of samples incorporated in training is relatively low, the resultant performance improvement is not markedly noticeable compared to the scenario where only timestamp labels are utilized.
	\begin{table}[t]
		\centering
		\caption{\sc The Performance of Our Model Trained with Mixed Supervision }
		\begin{tabular}{c|cccccc}
			\toprule
			\textbf{PCT.} & \textbf{0\%(TS)} & \textbf{20\%}  & \textbf{40\%}  &\textbf{60\%} &\textbf{80\%}  & \textbf{100\%} \\
			\midrule
			$F_m$  & 91.45\%  & 91.93\%  & 92.73\%  & 93.48\%  & 94.13\%  & 95.06\% \\

			\bottomrule             
		\end{tabular}
		\label{mixed-supervision}
	\end{table} 
	
	\section{Conclusion}
	In this paper, we proposed a novel weakly supervised method for wearable-based human activity segmentation and recognition that significantly improves performance compared to state-of-the-art methods. To reduce the quantity of training data annotations without compromising the model's performance, our approach relies on a multi-stage architecture supervised by timestamp annotations for joint activity segmentation and recognition. In our proposed method, we employ a sample-to-prototype contrast and an efficient pseudo-label generation strategy with contrastive learning and optimal transport theory to provide sample-level supervision for these unlabeled samples. Based on the four wearable-based HAR datasets, extensive experimental results proved that our model trained with timestamp labels achieves comparable performance to the fully supervised methods. Furthermore, leveraging our uniquely designed modules, our method shows a remarkable improvement in the class average F-score on the Hospital/Opportunity/PAMAP2/Skoda datasets by 10.45\%/17.68\%/38.84\%/37.33\% respectively, compared to the method with solely using timestamp labels. Compared with other state-of-the-art timestamp-supervised algorithms, our method improves the recognition performance ($F_m$) by 5.28\%/2.43\%/7.7\%/6.52\%, and the segmentation performance ($JI$) has a 6.53\%/2.28\%/7.42\%/14.48\% improvement on these four datasets.
	
	In future work, we will explore different weak supervision implementation methods, which can further improve the segmentation recognition performance of the model under the premise of reducing the labeling workload and applying it to more realistic application scenarios.

	\bibliography{ref}

\begin{thebibliography}{10}
\providecommand{\url}[1]{#1}
\csname url@samestyle\endcsname
\providecommand{\newblock}{\relax}
\providecommand{\bibinfo}[2]{#2}
\providecommand{\BIBentrySTDinterwordspacing}{\spaceskip=0pt\relax}
\providecommand{\BIBentryALTinterwordstretchfactor}{4}
\providecommand{\BIBentryALTinterwordspacing}{\spaceskip=\fontdimen2\font plus
\BIBentryALTinterwordstretchfactor\fontdimen3\font minus
  \fontdimen4\font\relax}
\providecommand{\BIBforeignlanguage}[2]{{%
\expandafter\ifx\csname l@#1\endcsname\relax
\typeout{** WARNING: IEEEtran.bst: No hyphenation pattern has been}%
\typeout{** loaded for the language `#1'. Using the pattern for}%
\typeout{** the default language instead.}%
\else
\language=\csname l@#1\endcsname
\fi
#2}}
\providecommand{\BIBdecl}{\relax}
\BIBdecl

\bibitem{xiao2022self}
C.~Xiao, S.~Chen, F.~Zhou, and J.~Wu, ``Self-supervised few-shot time-series
  segmentation for activity recognition,'' \emph{IEEE Transactions on Mobile
  Computing}, 2022.

\bibitem{hao2021invariant}
Y.~Hao, R.~Zheng, and B.~Wang, ``Invariant feature learning for sensor-based
  human activity recognition,'' \emph{IEEE Trans. Mobile Comput.}, 2021.

\bibitem{chen2021deep}
K.~Chen, D.~Zhang, L.~Yao, and et~al., ``Deep learning for sensor-based human
  activity recognition: Overview, challenges, and opportunities,'' \emph{ACM
  Comput. Surv.}, vol.~54, no.~4, pp. 1--40, 2021.

\bibitem{yu2021fedhar}
H.~Yu, Z.~Chen, X.~Zhang, X.~Chen, F.~Zhuang, H.~Xiong, and X.~Cheng, ``Fedhar:
  Semi-supervised online learning for personalized federated human activity
  recognition,'' \emph{IEEE Trans. Mobile Comput.}, 2021.

\bibitem{lv2018bi}
M.~Lv, L.~Chen, T.~Chen, and G.~Chen, ``Bi-view semi-supervised learning based
  semantic human activity recognition using accelerometers,'' \emph{IEEE
  Transactions on Mobile Computing}, vol.~17, no.~9, pp. 1991--2001, 2018.

\bibitem{pei2020mars}
L.~Pei, S.~Xia, L.~Chu, and et~al., ``Mars: Mixed virtual and real wearable
  sensors for human activity recognition with multi-domain deep learning
  model,'' \emph{IEEE Internet Things J.}, pp. 1--1, 2021.

\bibitem{yao2017deepsense}
S.~Yao, S.~Hu, Y.~Zhao, A.~Zhang, and T.~Abdelzaher, ``Deepsense: A unified
  deep learning framework for time-series mobile sensing data processing,'' in
  \emph{Int. World Wide Web Conf. (WWW)}, 2017, pp. 351--360.

\bibitem{yao2018efficient}
R.~Yao, G.~Lin, Q.~Shi, and D.~C. Ranasinghe, ``Efficient dense labelling of
  human activity sequences from wearables using fully convolutional networks,''
  \emph{Pattern Recognit.}, vol.~78, pp. 252--266, 2018.

\bibitem{9772403}
S.~Xia, L.~Chu, L.~Pei, W.~Yu, and R.~C. Qiu, ``A boundary consistency-aware
  multitask learning framework for joint activity segmentation and recognition
  with wearable sensors,'' \emph{IEEE Transactions on Industrial Informatics},
  vol.~19, no.~3, pp. 2984--2996, 2023.

\bibitem{ma2019attnsense}
H.~Ma, W.~Li, X.~Zhang, S.~Gao, and S.~Lu, ``Attnsense: Multi-level attention
  mechanism for multimodal human activity recognition.'' in \emph{Int. Joint
  Conf. Artif. Intell. (IJCAI)}, 2019, pp. 3109--3115.

\bibitem{qian2021weakly}
H.~Qian, S.~J. Pan, and C.~Miao, ``Weakly-supervised sensor-based activity
  segmentation and recognition via learning from distributions,'' \emph{Artif.
  Intell.}, vol. 292, p. 103429, 2021.

\bibitem{deryck2021change}
T.~Deryck, M.~De~Vos, and A.~Bertrand, ``Change point detection in time series
  data using autoencoders with a time-invariant representation,'' \emph{IEEE
  Trans. Signal Process.}, 2021.

\bibitem{tonekaboni2021unsupervised}
S.~Tonekaboni, D.~Eytan, and A.~Goldenberg, ``Unsupervised representation
  learning for time series with temporal neighborhood coding,'' \emph{ICLR
  2021}, 2021.

\bibitem{truong2020selective}
C.~Truong, L.~Oudre, and N.~Vayatis, ``Selective review of offline change point
  detection methods,'' \emph{Signal Processing}, vol. 167, p. 107299, 2020.

\bibitem{xiao2020deepseg}
C.~Xiao, Y.~Lei, Y.~Ma, F.~Zhou, and Z.~Qin, ``Deepseg: Deep-learning-based
  activity segmentation framework for activity recognition using wifi,''
  \emph{IEEE Internet of Things Journal}, vol.~8, no.~7, pp. 5669--5681, 2020.

\bibitem{zhou2022regional}
T.~Zhou, M.~Zhang, F.~Zhao, and J.~Li, ``Regional semantic contrast and
  aggregation for weakly supervised semantic segmentation,'' in
  \emph{Proceedings of the IEEE/CVF Conference on Computer Vision and Pattern
  Recognition}, 2022, pp. 4299--4309.

\bibitem{du2022weakly}
Y.~Du, Z.~Fu, Q.~Liu, and Y.~Wang, ``Weakly supervised semantic segmentation by
  pixel-to-prototype contrast,'' in \emph{Proceedings of the IEEE/CVF
  Conference on Computer Vision and Pattern Recognition}, 2022, pp. 4320--4329.

\bibitem{souri2021fast}
Y.~Souri, M.~Fayyaz, L.~Minciullo, G.~Francesca, and J.~Gall, ``Fast weakly
  supervised action segmentation using mutual consistency,'' \emph{IEEE Trans.
  Pattern Anal. Mach. Intell.}, 2021.

\bibitem{sayed2023new}
S.~Sayed, R.~Ghoddoosian, B.~Trivedi, and V.~Athitsos, ``A new dataset and
  approach for timestamp supervised action segmentation using human object
  interaction,'' in \emph{Proceedings of the IEEE/CVF Conference on Computer
  Vision and Pattern Recognition}, 2023, pp. 3132--3141.

\bibitem{rahaman2022generalized}
R.~Rahaman, D.~Singhania, A.~Thiery, and A.~Yao, ``A generalized and robust
  framework for timestamp supervision in temporal action segmentation,'' in
  \emph{Computer Vision--ECCV 2022: 17th European Conference, Tel Aviv, Israel,
  October 23--27, 2022, Proceedings, Part IV}.\hskip 1em plus 0.5em minus
  0.4em\relax Springer, 2022, pp. 279--296.

\bibitem{khan2022timestamp}
H.~Khan, S.~Haresh, A.~Ahmed, S.~Siddiqui, A.~Konin, M.~Z. Zia, and Q.-H. Tran,
  ``Timestamp-supervised action segmentation with graph convolutional
  networks,'' in \emph{2022 IEEE/RSJ International Conference on Intelligent
  Robots and Systems (IROS)}.\hskip 1em plus 0.5em minus 0.4em\relax IEEE,
  2022, pp. 10\,619--10\,626.

\bibitem{li2021temporal}
Z.~Li, Y.~Abu~Farha, and J.~Gall, ``Temporal action segmentation from timestamp
  supervision,'' in \emph{Proceedings of the IEEE/CVF Conference on Computer
  Vision and Pattern Recognition}, 2021, pp. 8365--8374.

\bibitem{xia2021learning}
S.~Xia, L.~Chu, L.~Pei, and et~al., ``Learning disentangled representation for
  mixed-reality human activity recognition with a single imu sensor,''
  \emph{IEEE Trans. Instrum. Meas.}, vol.~70, pp. 1--14, 2021.

\bibitem{radu2018multimodal}
V.~Radu, C.~Tong, S.~Bhattacharya, N.~D. Lane, C.~Mascolo, M.~K. Marina, and
  F.~Kawsar, ``Multimodal deep learning for activity and context recognition,''
  \emph{Proc. ACM Interact. Mob. Wearable Ubiquitous Technol. (IMWUT/UbiComp)},
  vol.~1, no.~4, pp. 1--27, 2018.

\bibitem{yang2015deep}
J.~Yang, M.~N. Nguyen, P.~P. San, X.~Li, and S.~Krishnaswamy, ``Deep
  convolutional neural networks on multichannel time series for human activity
  recognition.'' in \emph{Int. Joint Conf. Artif. Intell. (IJCAI)},
  vol.~15.\hskip 1em plus 0.5em minus 0.4em\relax Buenos Aires, Argentina,
  2015, pp. 3995--4001.

\bibitem{guan2017ensembles}
Y.~Guan and T.~Pl{\"o}tz, ``Ensembles of deep lstm learners for activity
  recognition using wearables,'' \emph{Proc. ACM Interact. Mob. Wearable
  Ubiquitous Technol. (IMWUT/UbiComp)}, vol.~1, no.~2, pp. 1--28, 2017.

\bibitem{chen2019semisupervised}
K.~Chen, L.~Yao, D.~Zhang, X.~Wang, X.~Chang, and F.~Nie, ``A semisupervised
  recurrent convolutional attention model for human activity recognition,''
  \emph{IEEE Trans. Neural Netw. Learn. Syst.}, vol.~31, no.~5, pp. 1747--1756,
  2019.

\bibitem{ordonez2016deep}
F.~J. Ord{\'o}{\~n}ez and D.~Roggen, ``Deep convolutional and lstm recurrent
  neural networks for multimodal wearable activity recognition,''
  \emph{Sensors}, vol.~16, no.~1, p. 115, 2016.

\bibitem{banos2015multiwindow}
O.~Banos, J.-M. Galvez, M.~Damas, A.~Guillen, L.-J. Herrera, H.~Pomares,
  I.~Rojas, C.~Villalonga, C.~S. Hong, and S.~Lee, ``Multiwindow fusion for
  wearable activity recognition,'' in \emph{Int. Conf. on Artif. Neur. Networks
  (IWANN)}.\hskip 1em plus 0.5em minus 0.4em\relax Springer, 2015, pp.
  290--297.

\bibitem{noor2017adaptive}
M.~H.~M. Noor, Z.~Salcic, and I.~e.~a. Kevin, ``Adaptive sliding window
  segmentation for physical activity recognition using a single tri-axial
  accelerometer,'' \emph{Pervasive Mob. Comput.}, vol.~38, pp. 41--59, 2017.

\bibitem{zhang2019human}
Y.~Zhang, Z.~Zhang, Y.~Zhang, and et~al., ``Human activity recognition based on
  motion sensor using u-net,'' \emph{IEEE Access}, vol.~7, pp.
  75\,213--75\,226, 2019.

\bibitem{zhang2021conditional}
L.~Zhang, W.~Zhang, and N.~Japkowicz, ``Conditional-unet: A condition-aware
  deep model for coherent human activity recognition from wearables,'' in
  \emph{Int. Conf. Pattern Recognit. (ICPR)}, 2021, pp. 5889--5896.

\bibitem{xia2022multi}
S.~Xia, L.~Chu, L.~Pei, W.~Yu, and R.~C. Qiu, ``Multi-level contrast network
  for wearables-based joint activity segmentation and recognition,'' in
  \emph{GLOBECOM 2022-2022 IEEE Global Communications Conference}.\hskip 1em
  plus 0.5em minus 0.4em\relax IEEE, 2022, pp. 566--572.

\bibitem{ding2022temporal}
G.~Ding, F.~Sener, and A.~Yao, ``Temporal action segmentation: An analysis of
  modern technique,'' \emph{arXiv preprint arXiv:2210.10352}, 2022.

\bibitem{farha2019ms}
Y.~A. Farha and J.~Gall, ``Ms-tcn: Multi-stage temporal convolutional network
  for action segmentation,'' in \emph{Proc. IEEE Conf. Comput. Vis. Pattern
  Recognit. (CVPR)}, 2019, pp. 3575--3584.

\bibitem{wang2020boundary}
Z.~Wang, Z.~Gao, L.~Wang, Z.~Li, and G.~Wu, ``Boundary-aware cascade networks
  for temporal action segmentation,'' in \emph{Eur. Conf. Comput. Vis.
  (ECCV)}.\hskip 1em plus 0.5em minus 0.4em\relax Springer, 2020, pp. 34--51.

\bibitem{ishikawa2021alleviating}
Y.~Ishikawa, S.~Kasai, Y.~Aoki, and H.~Kataoka, ``Alleviating over-segmentation
  errors by detecting action boundaries,'' in \emph{IEEE Winter Conf. Appl.
  Comput. Vis. (WACV)}, 2021, pp. 2322--2331.

\bibitem{yi2021asformer}
F.~Yi, H.~Wen, and T.~Jiang, ``Asformer: Transformer for action segmentation,''
  \emph{British Machine Vision Conference}, 2021.

\bibitem{bojanowski2014weakly}
P.~Bojanowski, R.~Lajugie, F.~Bach, I.~Laptev, J.~Ponce, C.~Schmid, and
  J.~Sivic, ``Weakly supervised action labeling in videos under ordering
  constraints,'' in \emph{Computer Vision--ECCV 2014: 13th European Conference,
  Zurich, Switzerland, September 6-12, 2014, Proceedings, Part V 13}.\hskip 1em
  plus 0.5em minus 0.4em\relax Springer, 2014, pp. 628--643.

\bibitem{behrmann2022unified}
N.~Behrmann, S.~A. Golestaneh, Z.~Kolter, J.~Gall, and M.~Noroozi, ``Unified
  fully and timestamp supervised temporal action segmentation via sequence to
  sequence translation,'' in \emph{Computer Vision--ECCV 2022: 17th European
  Conference, Tel Aviv, Israel, October 23--27, 2022, Proceedings, Part
  XXXV}.\hskip 1em plus 0.5em minus 0.4em\relax Springer, 2022, pp. 52--68.

\bibitem{richard2018neuralnetwork}
A.~Richard, H.~Kuehne, A.~Iqbal, and J.~Gall, ``Neuralnetwork-viterbi: A
  framework for weakly supervised video learning,'' in \emph{Proceedings of the
  IEEE conference on Computer Vision and Pattern Recognition}, 2018, pp.
  7386--7395.

\bibitem{ridley2022transformers}
J.~Ridley, H.~Coskun, D.~J. Tan, N.~Navab, and F.~Tombari, ``Transformers in
  action: weakly supervised action segmentation,'' \emph{arXiv preprint
  arXiv:2201.05675}, 2022.

\bibitem{oord2018representation}
A.~v.~d. Oord, Y.~Li, and O.~Vinyals, ``Representation learning with
  contrastive predictive coding,'' \emph{arXiv preprint arXiv:1807.03748},
  2018.

\bibitem{chen2020simple}
T.~Chen, S.~Kornblith, M.~Norouzi, and G.~Hinton, ``A simple framework for
  contrastive learning of visual representations,'' in \emph{Int. Conf. Mach.
  Learn. (ICML)}.\hskip 1em plus 0.5em minus 0.4em\relax PMLR, 2020, pp.
  1597--1607.

\bibitem{deldari2021time}
S.~Deldari, D.~V. Smith, H.~Xue, and F.~D. Salim, ``Time series change point
  detection with self-supervised contrastive predictive coding,'' in
  \emph{Proc. World Wide Web Conf. (WWW)}, 2021, pp. 3124--3135.

\bibitem{eldele2021time}
E.~Eldele, M.~Ragab, Z.~Chen, M.~Wu, C.~K. Kwoh, X.~Li, and C.~Guan,
  ``Time-series representation learning via temporal and contextual
  contrasting,'' \emph{Int. Joint Conf. Artif. Intell. (IJCAI 2021)}, 2021.

\bibitem{khosla2020supervised}
P.~Khosla, P.~Teterwak, C.~Wang, and et~al., ``Supervised contrastive
  learning,'' \emph{Adv. neural inf. proces. syst. (NeurIPS)}, vol.~33, pp.
  18\,661--18\,673, 2020.

\bibitem{wang2021exploring}
W.~Wang, T.~Zhou, F.~Yu, J.~Dai, E.~Konukoglu, and L.~Van~Gool, ``Exploring
  cross-image pixel contrast for semantic segmentation,'' in \emph{Proc. IEEE
  Int. Conf. Comput. Vision (ICCV)}, 2021, pp. 7303--7313.

\bibitem{su2017order}
B.~Su and G.~Hua, ``Order-preserving wasserstein distance for sequence
  matching,'' in \emph{Proceedings of the IEEE conference on computer vision
  and pattern recognition}, 2017, pp. 1049--1057.

\bibitem{lim2022order}
Y.~C.~F. Lim, L.~Wynter, and S.~H. Lim, ``Order constraints in optimal
  transport,'' in \emph{International Conference on Machine Learning}.\hskip
  1em plus 0.5em minus 0.4em\relax PMLR, 2022, pp. 13\,313--13\,333.

\bibitem{zhou2016learning}
B.~Zhou, A.~Khosla, A.~Lapedriza, A.~Oliva, and A.~Torralba, ``Learning deep
  features for discriminative localization,'' in \emph{Proceedings of the IEEE
  conference on computer vision and pattern recognition}, 2016, pp. 2921--2929.

\bibitem{chen2020weakly}
L.~Chen, W.~Wu, C.~Fu, X.~Han, and Y.~Zhang, ``Weakly supervised semantic
  segmentation with boundary exploration,'' in \emph{Computer Vision--ECCV
  2020: 16th European Conference, Glasgow, UK, August 23--28, 2020,
  Proceedings, Part XXVI 16}.\hskip 1em plus 0.5em minus 0.4em\relax Springer,
  2020, pp. 347--362.

\bibitem{kalantidis2020hard}
Y.~Kalantidis, M.~B. Sariyildiz, N.~Pion, P.~Weinzaepfel, and D.~Larlus, ``Hard
  negative mixing for contrastive learning,'' \emph{Advances in Neural
  Information Processing Systems}, vol.~33, pp. 21\,798--21\,809, 2020.

\bibitem{kumar2021unsupervised}
S.~Kumar, S.~Haresh, A.~Ahmed, and et~al., ``Unsupervised activity segmentation
  by joint representation learning and online clustering,'' \emph{Proc. IEEE
  Conf. Comput. Vis. Pattern Recognit. (CVPR)}, 2022.

\bibitem{gammulle2021tmmf}
H.~Gammulle, S.~Denman, S.~Sridharan, and C.~Fookes, ``Tmmf: Temporal
  multi-modal fusion for single-stage continuous gesture recognition,''
  \emph{IEEE Trans. Image Process.}, vol.~30, pp. 7689--7701, 2021.

\bibitem{abedin2021attend}
A.~Abedin, M.~Ehsanpour, Q.~Shi, H.~Rezatofighi, and D.~C. Ranasinghe, ``Attend
  and discriminate: Beyond the state-of-the-art for human activity recognition
  using wearable sensors,'' \emph{Proc. ACM Interact. Mob. Wearable Ubiquitous
  Technol. (IMWUT/UbiComp)}, vol.~5, no.~1, pp. 1--22, 2021.

\bibitem{ward2011performance}
J.~A. Ward, P.~Lukowicz, and H.~W. Gellersen, ``Performance metrics for
  activity recognition,'' \emph{ACM Trans. Intell. Sys. Tech. (TIST)}, vol.~2,
  no.~1, pp. 1--23, 2011.

\bibitem{chavarriaga2013opportunity}
R.~Chavarriaga, H.~Sagha, A.~Calatroni, S.~T. Digumarti, G.~Tr{\"o}ster,
  J.~d.~R. Mill{\'a}n, and D.~Roggen, ``The opportunity challenge: A benchmark
  database for on-body sensor-based activity recognition,'' \emph{Pattern
  Recognit. Lett.}, vol.~34, no.~15, pp. 2033--2042, 2013.

\bibitem{reiss2012introducing}
A.~Reiss and D.~Stricker, ``Introducing a new benchmarked dataset for activity
  monitoring,'' in \emph{Proc. Int. Symp. Wearable Comput. (ISWC)}.\hskip 1em
  plus 0.5em minus 0.4em\relax IEEE, 2012, pp. 108--109.

\bibitem{stiefmeier2008wearable}
T.~Stiefmeier, D.~Roggen, G.~Ogris, P.~Lukowicz, and G.~Tr{\"o}ster, ``Wearable
  activity tracking in car manufacturing,'' \emph{IEEE Pervasive Computing},
  vol.~7, no.~2, pp. 42--50, 2008.

\bibitem{murahari2018attention}
V.~S. Murahari and T.~Pl{\"o}tz, ``On attention models for human activity
  recognition,'' in \emph{Proc. ACM Int. Symp. Wearable Comput. (ISWC)}, 2018,
  pp. 100--103.

\end{thebibliography}

	
	%

	

	\ifCLASSOPTIONcompsoc

	\ifCLASSOPTIONcaptionsoff
	\newpage
	\fi

	\begin{IEEEbiography}[{\includegraphics[width=1in,height=1.25in,clip,keepaspectratio]{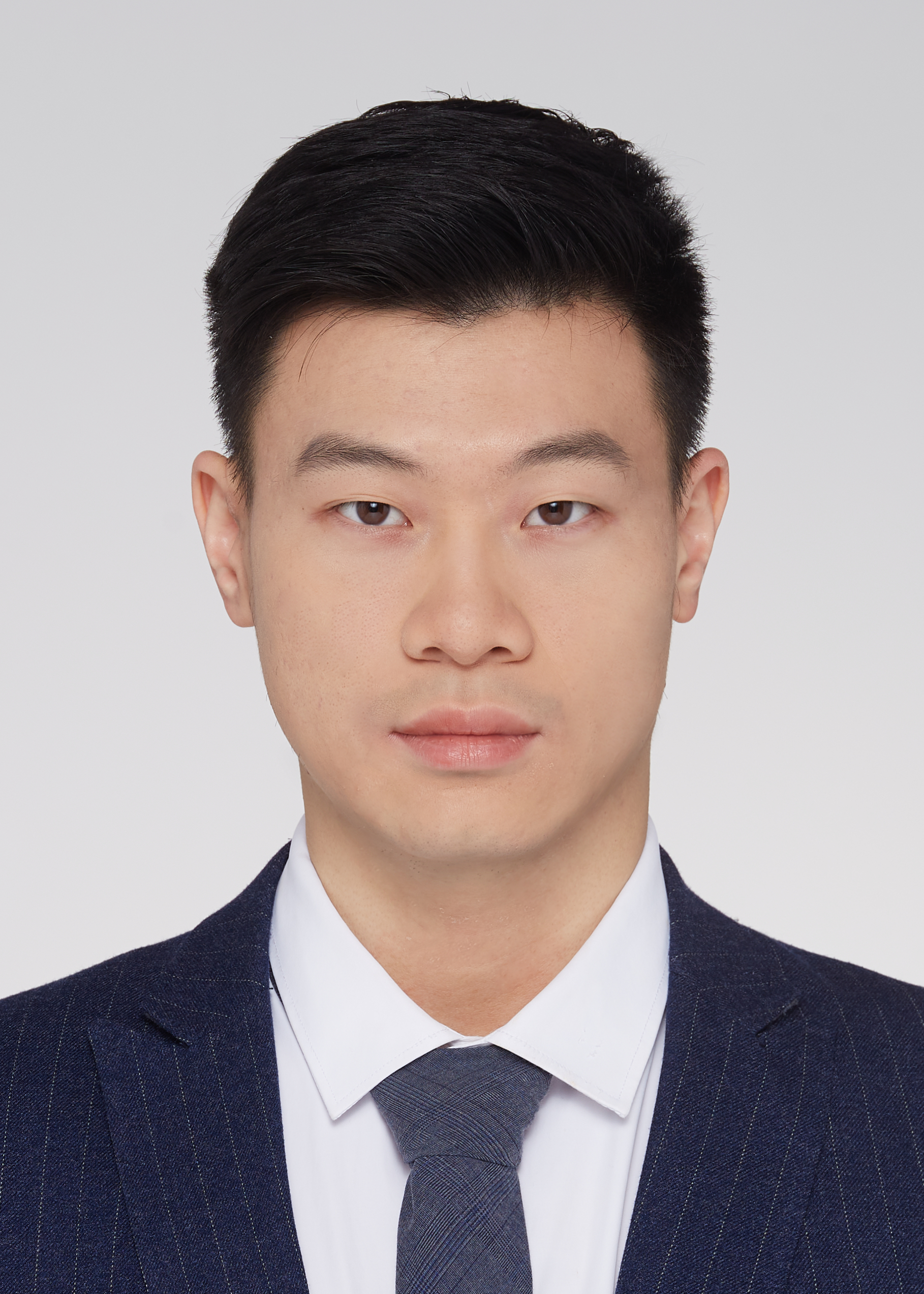}}]{Songpengcheng Xia}
		(Student Member. IEEE) received the B.S. degree in navigation engineering from Wuhan University, Wuhan, China, in 2019. He is currently pursuing the Ph.D. degree with Shanghai Jiao Tong University, Shanghai, China. His current research interests include machine learning, inertial navigation, multi-sensor fusion and wearable-based human activity recognition.
	\end{IEEEbiography}
	
	\begin{IEEEbiography}[{\includegraphics[width=1in,height=1.25in,clip,keepaspectratio]{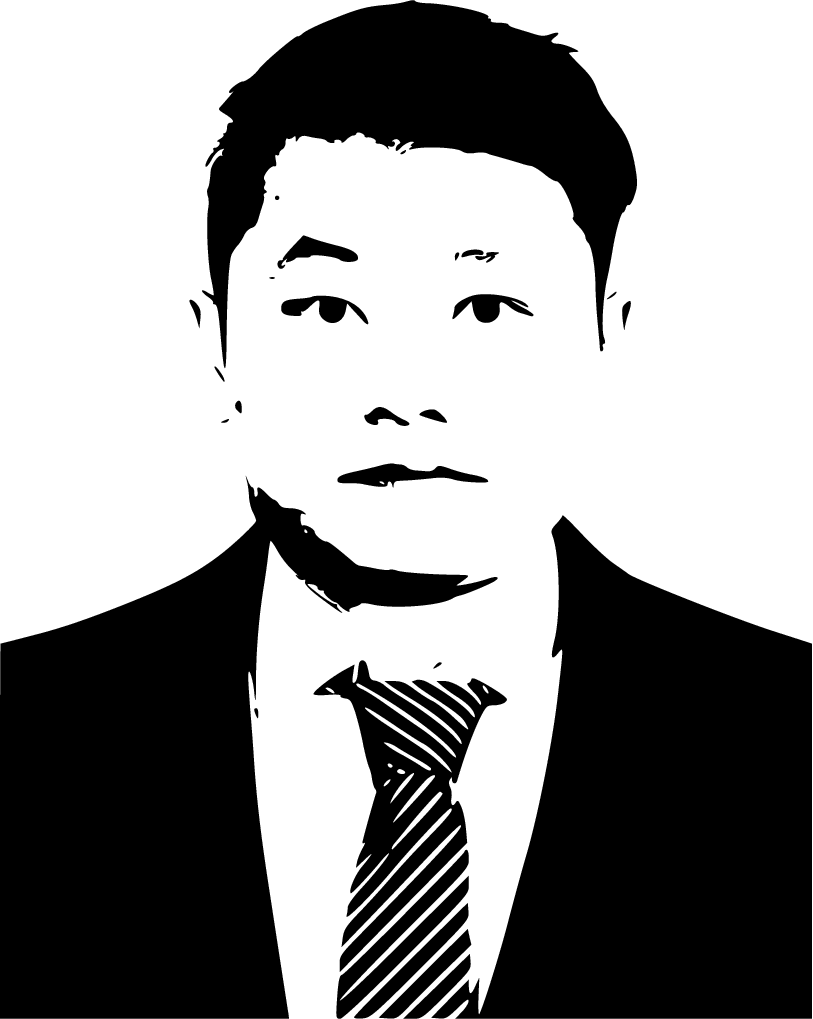}}]{Lei Chu}
		(Senior Member, IEEE) received the
		Ph.D. degree from Shanghai Jiao Tong University
		(SJTU) in December 2019. He has been a full-time
		Research Scholar with the University of Southern
		California, Los Angeles, CA, USA, hosted by
		Prof. Andreas F. Molisch, since 2021. Before that,
		he was a senior research associate with the School of Electronics, Information, and Electrical Engineering,
		SJTU. He was a Visiting Scholar with the University
		of Tennessee, Knoxville, in 2019. He has contributed
		to three book chapters, authored over fifty papers in refereed
		journals/conferences, and issued several patents. His current research interests
		include integrating information theory into neural network optimization and
		extending them into B5G wireless communications and intelligent sensing
		applications. He serves as a regular reviewer for dozens of IEEE journals and
		conferences. He is an associate/academic editor for \textit{IET Electrical Systems in Transportation, Hindawi Journal of Sensors, Journal of Artificial Intelligence and Big Data}.
	\end{IEEEbiography}
	
	\begin{IEEEbiography}[{\includegraphics[width=1in,height=1.25in,clip,keepaspectratio]{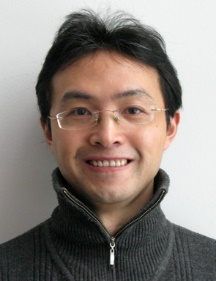}}]{Ling Pei}
		(Senior Member, IEEE) received the Ph.D. degree from Southeast University, Nanjing, China, in 2007. From 2007 to 2013, he was a Specialist Research Scientist with the Finnish Geospatial Research Institute. He is currently a Professor at the School of Electronic Information and Electrical Engineering, Shanghai Jiao Tong University. He has authored or co-authored over 100 scientific papers. He is also an inventor of 25 patents and pending patents. His main research is in the areas of indoor/outdoor seamless positioning, ubiquitous computing, wireless positioning, Bio-inspired navigation, context-aware applications, location-based services, and navigation of unmanned systems. Dr. Pei was a recipient of the Shanghai Pujiang Talent in 2014 and ranked as the World's Top 2\% scientists by Stanford University in 2022.
	\end{IEEEbiography}

        \begin{IEEEbiography}[{\includegraphics[width=1in,height=1.25in,clip,keepaspectratio]{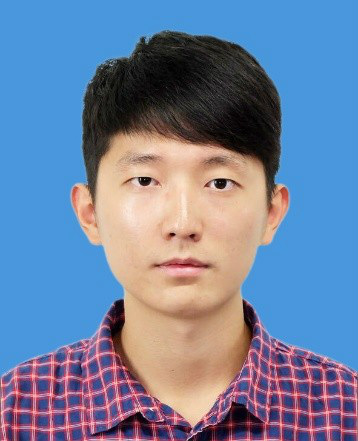}}]{Jiarui Yang}
		(Student Member, IEEE) received the B.S. degree in telecommunication engineering from Politecnico di Torino, Turin, Italy in 2018. He received the M.S. degree in communication systems from KTH Royal Institute of Technology, Stockholm, Sweden in 2021. He is currently working toward the Ph.D. degree in information and communication engineering with Shanghai Jiao Tong University, Shanghai, China. His research interests include machine learning, deep learning, multi-sensor fusion, and human pose estimation.

	\end{IEEEbiography}
	
	\begin{IEEEbiography}[{\includegraphics[width=1in,height=1.25in,clip,keepaspectratio]{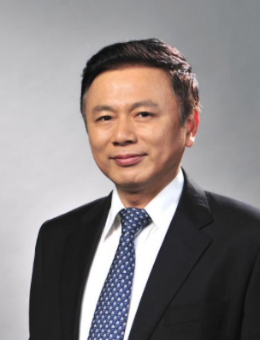}}]{Wenxian Yu}
		(Senior Member, IEEE) received the B.S., M.S., and Ph.D. degrees from the National University of Defense Technology, Changsha, China, in 1985, 1988, and 1993, respectively.
		From 1996 to 2008, he was a Professor with the College of Electronic Science and Engineering, National University of Defense Technology, where he was also the Deputy Head of the College and an Assistant Director of the National Key Laboratory of Automatic Target Recognition. From 2009 to 2011, he was the Executive Dean of the School of Electronic, Information, and Electrical Engineering, Shanghai Jiao Tong University, Shanghai, China. He is currently a Yangtze River Scholar Distinguished Professor and the Head of the research part in the School of Electronic, Information,
		and Electrical Engineering, Shanghai Jiao Tong University. His research interests include remote sensing information processing, automatic target recognition, multisensor data fusion, etc.
	\end{IEEEbiography}
	
	\begin{IEEEbiography}[{\includegraphics[width=1in,height=1.25in,clip,keepaspectratio]{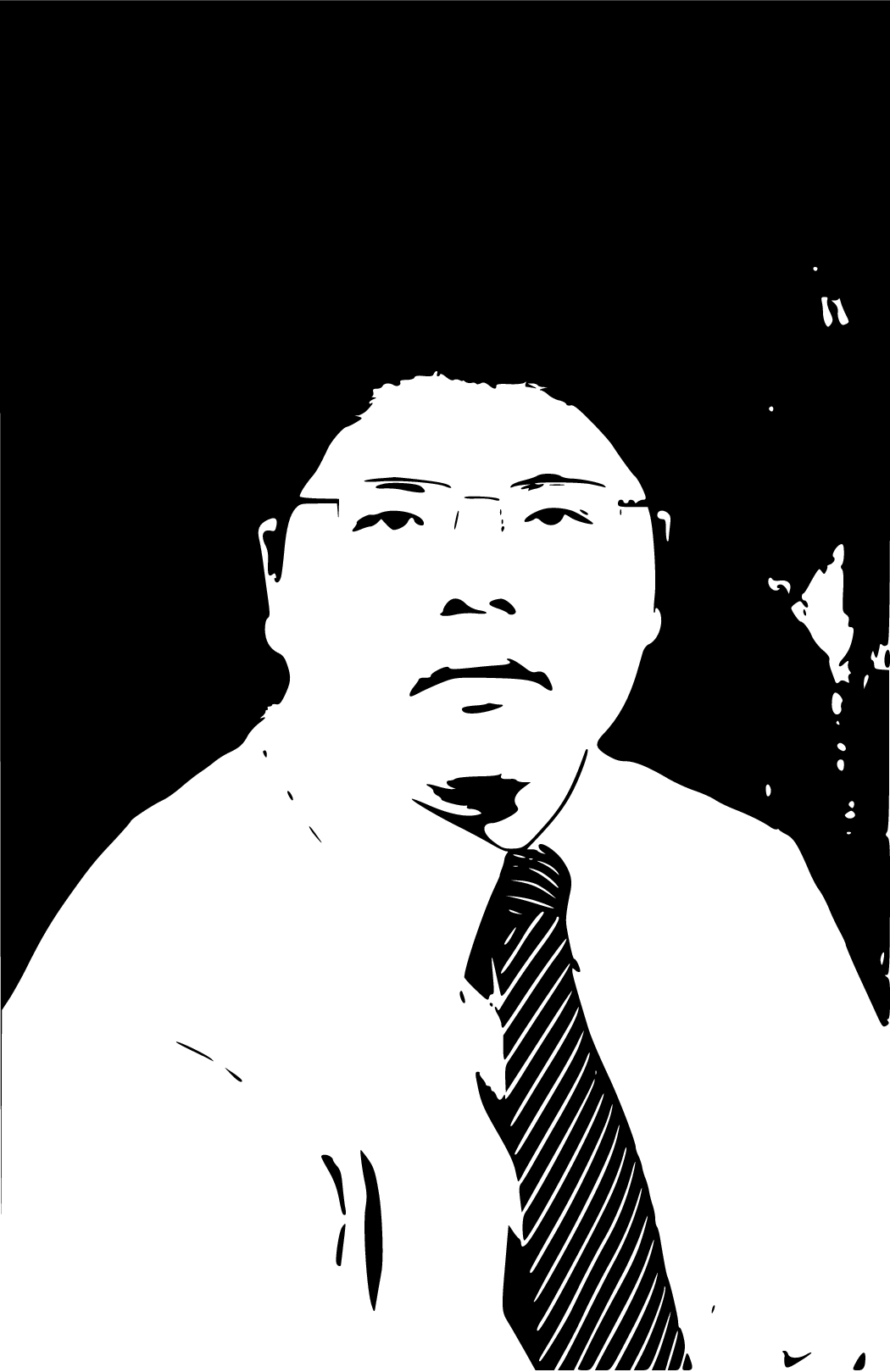}}]{Robert C. Qiu}
		(IEEE S'93-M'96-SM'01-F'14) received the Ph.D.
		degree in electrical engineering from New York University. He joined the School of Electronic Information and Communications, Huazhong University of
		Science and Technology, China, as a full Professor
		since 2020.
		He was Founder-CEO and President of Wiscom
		Technologies, Inc., manufacturing and marketing
		WCDMA chipsets. Wiscom was sold to Intel in
		2003. Before Wiscom, he worked for GTE Labs,
		Inc. (now Verizon), Waltham, MA, and Bell Labs, Lucent, Whippany, NJ.
		He has worked in wireless communications and network, machine learning,
		Smart Grid, digital signal processing, EM scattering, composite absorbing
		materials, RF microelectronics, UWB, underwater acoustics, and fiber optics.
		He holds dozens of patents and authored over 100 journal papers and 100 conference papers. He has 15 contributions to 3GPP and IEEE
		standards bodies. In 1998 he developed the first three courses on 3G for
		Bell Labs researchers. He started as an Associate Professor in 2003 in the
		Department of Electrical and Computer Engineering, Center for Manufacturing Research, Tennessee Technological University, Cookeville, Tennessee,
		where he became a Professor in 2008. He served as an adjunct professor in
		Polytechnic University, Brooklyn, New York. He was
		elected a Fellow of IEEE
		in 2015 for his contributions to ultra-wideband wireless communications.
		His current interests are random
		matrix theory based theoretical analysis for deep learning and its applications.
		
	\end{IEEEbiography}
	
	
	
	

\end{document}